\documentclass[10pt,twocolumn,letterpaper]{article}

\usepackage{cvpr}              

\usepackage{graphicx}
\usepackage{amsmath}
\usepackage{amssymb}
\usepackage{booktabs}
\usepackage{enumitem}
\usepackage[accsupp]{axessibility}  

\usepackage{makecell}

\usepackage{multirow}
\usepackage{comment}
\usepackage{textcomp}

\usepackage[dvipsnames]{xcolor}

\newcommand{\mathdelta}[1]{{\color{NavyBlue} ($#1$)}}
\newcommand{\mathdeltaneg}[1]{{\color{Red} ($#1$)}}

\usepackage[pagebackref,breaklinks,colorlinks]{hyperref}

\newcommand{\Scratch}{{From scratch}\xspace}
\newcommand{\ImNet}{{ImageNet}\xspace}
\newcommand{\JFT}{{JFT-300M}\xspace}
\newcommand{\MSCOCO}{{MSCOCO}\xspace}
\newcommand{\LVIS}{{LVIS}\xspace}
\newcommand{\Imagenet}{{ImageNet}\xspace}
\newcommand{\nasfpn}{{NAS-FPN}\xspace}
\newcommand{\fpn}{{FPN}\xspace}
\newcommand{\cascade}{{Cascade}\xspace}

\newcommand{\resnet}{{ResNet}\xspace}

\newcommand{\efficientNet}{{EfficientNet-B7}\xspace}
\newcommand{\frcnn}{{Fast-RCNN}\xspace}
\newcommand{\mrcnn}{{Mask-RCNN}\xspace}
\newcommand{\denselist}{\itemsep 0pt\topsep-6pt\partopsep-6pt}

\usepackage[capitalize]{cleveref}
\crefname{section}{Sec.}{Secs.}
\Crefname{section}{Section}{Sections}
\Crefname{table}{Table}{Tables}
\crefname{table}{Tab.}{Tabs.}

\def\cvprPaperID{7010} 
\def\confName{CVPR}
\def\confYear{2022}

\begin{document}

\title{Proper Reuse of Image Classification Features Improves Object Detection}

\author{Cristina Vasconcelos, Vighnesh Birodkar, Vincent Dumoulin\\
Google Research\\
{\tt\small \{crisnv,vighneshb,vdumoulin\}@google.com}
}
\twocolumn[{
\maketitle
    \begin{center}
        \captionsetup{type=figure}
        \begin{minipage}{.45\textwidth}
            \centering
            \includegraphics[width=\textwidth]{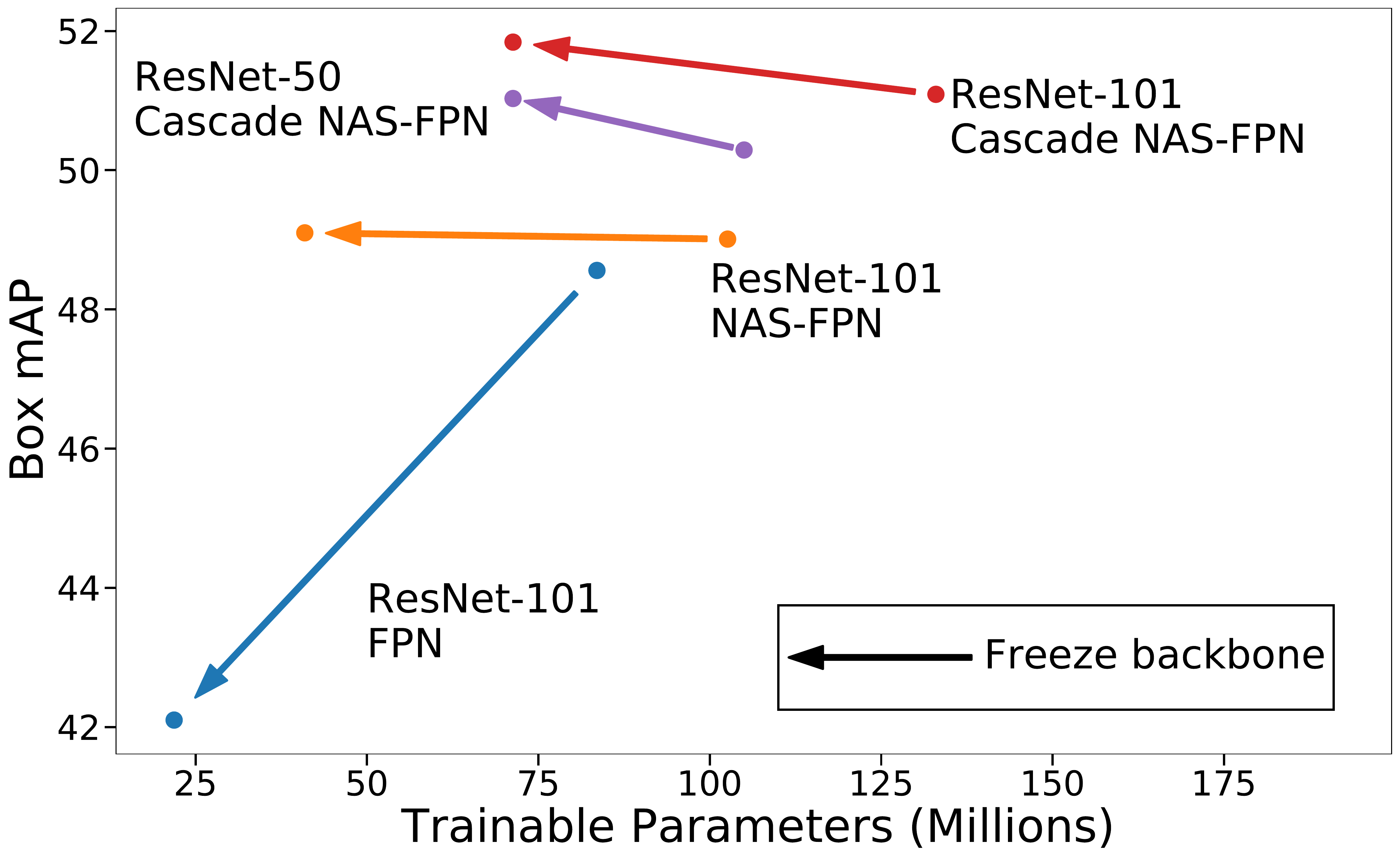}
        \end{minipage}%
        \hspace{15pt}
        \begin{minipage}{0.45\textwidth}
            \centering
            \includegraphics[width=\textwidth]{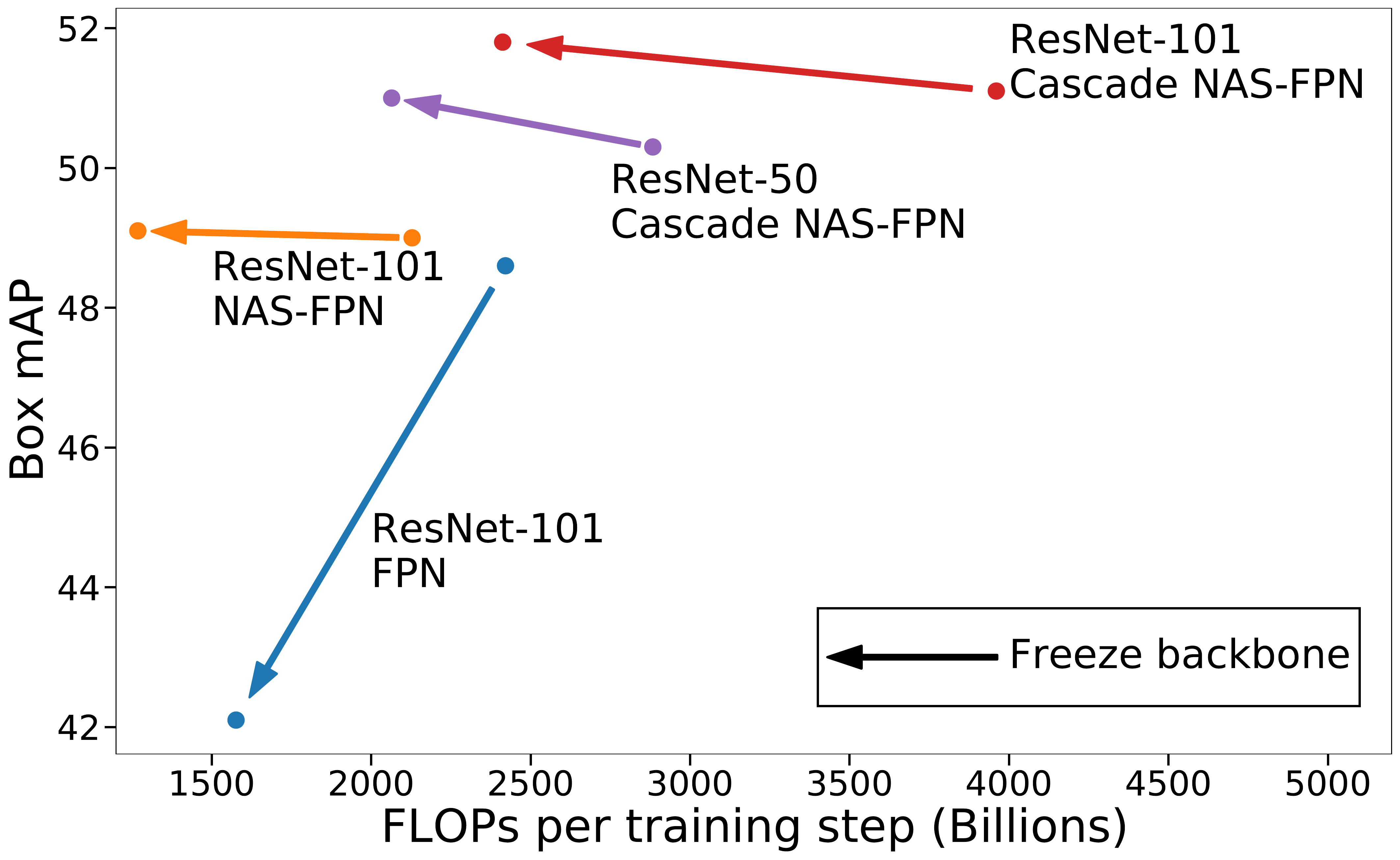}
        \end{minipage}
        \captionof{figure}{\label{fig:parameter_analysis}
        Impact on object detection performance of preserving and not fine-tuning features learned on \Imagenet. 
        The arrows indicate impact on each model, when trained with frozen backbone weights. 
        As long as the remaining detector components have enough capacity {\bf(left)}, 
        freezing increases performance while significantly reducing resources used during training ({\bf right}).}
    \end{center}
}]

\begin{abstract}
A common practice in transfer learning is to initialize the downstream model weights by pre-training on a data-abundant upstream task. In object detection specifically, the feature backbone is typically initialized with \Imagenet classifier weights and fine-tuned on the object detection task. Recent works show this is not strictly necessary under longer training regimes and provide recipes for training the backbone from scratch. We investigate the opposite direction of this end-to-end training trend: we show that an extreme form of knowledge preservation---freezing the classifier-initialized backbone---
consistently improves many different detection models, 
and leads to considerable resource savings.
We hypothesize and corroborate experimentally that the remaining detector components capacity and structure is a crucial factor in leveraging the frozen backbone. Immediate applications of our findings include performance improvements on hard cases like detection of long-tail object classes and computational and memory resource savings that contribute to
making the field more accessible to researchers with 
access to fewer computational resources.
\end{abstract}

\section{Introduction}
\label{sec:intro}

Transfer learning~\cite{pan2009survey} is a widely adopted practice in deep learning, especially when the target task has a smaller dataset;
the model is first pre-trained on an upstream task 
in which a larger amount data is available and then fine-tuned on the target task. Transfer learning from \ImNet or even larger~\cite{hinton2015distilling,chollet2017xception,SunShr17} or weakly labeled~\cite{mahajan2018exploring} datasets was  repeatedly shown to yield performance improvements across various vision tasks, architectures, and training procedures~\cite{SunShr17,mahajan2018exploring,Kornblith_2019_CVPR}. 

\begin{figure}
    \centering
    \includegraphics[width=0.6\linewidth]{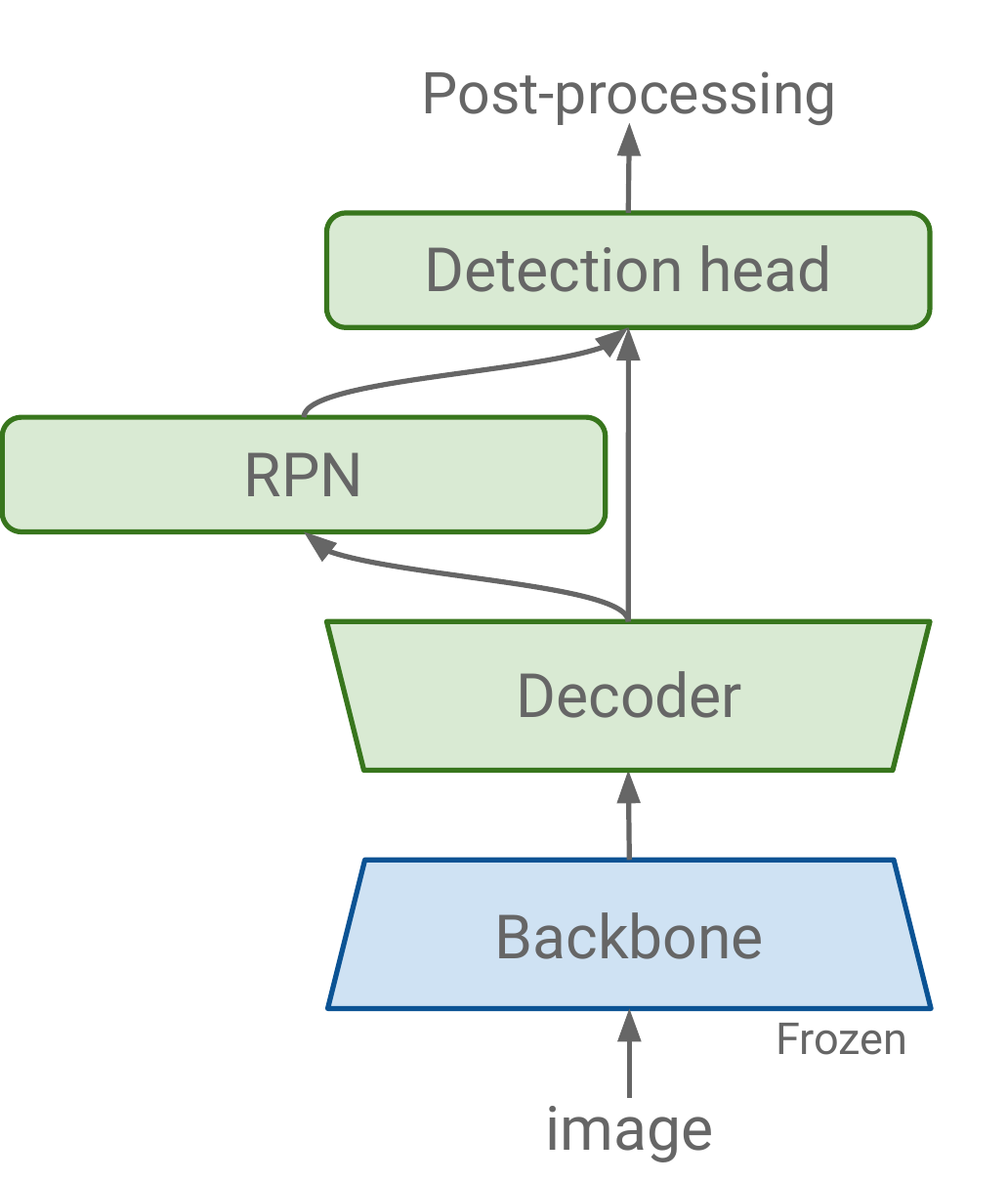}
    \caption{\label{fig:object_detection_diagram}A sketch of the Faster-RCNN detection model that serves
    as an example.
    Our proposed training procedure freezes the backbone
    after initializing it from a classification task and trains
    the other components from scratch.}
\end{figure}

For object detection\cite{od_survey} it is common practice to initialize the model's backbone (see \autoref{fig:object_detection_diagram} for a model diagram) with weight values obtained by pretraining on an image classification task, such as \ImNet~\cite{russakovsky2015imagenet}. Traditionally, the backbone is fine-tuned while training the other detector components from scratch. Two lines of work have recently made seemingly contradictory observations on transfer learning for object detection. On one hand, Sun et al.~\cite{SunShr17} show that object detectors benefit from the amount of classification data
used in pre-training. 
On the other hand, more recent papers have reported that the performance gap between transferring from a pre-trained backbone initialization and training the backbone from scratch with smaller, in-domain datasets vanishes with longer training~\cite{He2019RethinkingIP,Shen2020,li2021rethinking,du2021simple}.

We revisit transfer learning in its simplest form, where the backbone's classifier initialization is frozen during detection training.
This allows us to better understand the usefulness of the pre-trained representation without confounding factors resulting from fine-tuning. This approach has many advantages: it is simple, resource saving, and easy to replicate.
Moreover, using this approach, we are able to make the following two observations:
\begin{enumerate}[leftmargin=*]\denselist
    \item Longer training is a confounding factor in investigating the usefulness of the pre-trained representation, because the weights of a fine-tuned backbone move further away from their pre-trained initialization.
    \item This is important, because our ablation studies show that the representation learned on the upstream classification task is better for object detection than the one obtained from fine-tuning or training from scratch on the object detection task itself using a smaller in-domain dataset.
\end{enumerate}

When we preserve the pre-trained representation through freezing the backbone, we observe a consistent performance improvement from pre-training on larger datasets. Provided that the subsequent detector components have enough capacity, the models trained with a frozen backbone even exceed the performance of their fine-tuned or ``from scratch'' counterparts. 

As an immediate contribution of our findings, we show that it is possible to train an off-the-shelf object detection model with similar or superior performance while significantly reducing the need for computational resources, both memory-wise and computationally-wise (FLOPs)(\autoref{fig:parameter_analysis}). The performance benefits of the proposed upstream task knowledge preservation are even more clear when stratifying results by classes and the number of annotations available. Our results show that our extreme formulation of model reuse has a clear positive impact on classes with a low number of annotations, such as those found in long-tail object recognition.

\section{Related work}
\label{sec:related}

\noindent{\bf Benefits of object detector backbone pre-training.} The hypothesis that it is possible to improve performance on vision tasks by training a better base model has been largely investigated~\cite{Kornblith_2019_CVPR,fasterrcnn,retinanet}. Huang et al.~\cite{Huang16} focus on the close correlation between architectures with different capacity, their classification performances on \ImNet and their corresponding object detection performance. Sun et al.~\cite{SunShr17} shifts the discussion away from model comparison to focus on the impact of pre-training data. Their ablations contrast backbone pretraining on \ImNet versus \JFT classification tasks to corroborate the hypothesis that large-scale data helps in representation learning that ultimately improves transfer learning performance in classification, detection, segmentation, and pose estimation tasks. Importantly for our discussion, they point out that the benefit of pre-training from a larger dataset is capacity bounded. 

\medskip\noindent{\bf Unsupervised, weakly-supervised and self-supervised pre-training.} Reducing the need for task-specific annotations is especially relevant for tasks requiring fine-grained annotations such as detection and segmentation. This motivates an active area of research on unsupervised~\cite{Dai_2021_CVPR,bar2021detreg}, self-supervised  \cite{Rosenberg2005, Jeong_2019, sohn2020detection, xu2021end} and weakly supervised methods for object detection (pre-)training~\cite{Liang2015,Mahajan2018ExploringTL, predet, scaling_od_classification_weights}. 
In a  self-training formulation, Barret et al. \cite{Barret2020} propose to discard the original labels from \ImNet and obtain pseudo-labels using a detector trained on \MSCOCO. The pseudo-labeled ImageNet and labeled COCO data are then combined to train a new model. Closer to our work, Dhruv et al. \cite{Mahajan2018ExploringTL} investigate pre-training  on billions of weakly-labeled images using social media hashtags. They show that when using large amounts of pre-training data, detection performance is bounded by model capacity. Gains on smaller models are small or negative, but as model capacity increases the larger pre-training datasets yield consistent improvements. On the impact of pre-training using noisy labels, they conjecture that gains from weakly supervised pre-training compared to supervised pre-training (\ImNet) may be primarily due to improved object classification performance, rather than spatial localization performance. The contrast between gains in AP versus AP\@50 is used as a proxy for this analysis.\footnote{AP measures the average precision at different intersection over union (IoU) thresholds ranging from 0.5:0.95, while AP@50 is computed as the average precision computed at IoU threshold 0.5 only.}

\medskip\noindent{\bf Training from scratch.} He et al.~\cite{He2019RethinkingIP} show that it is possible to train an object detector from scratch from random initialization using in-domain supervision only but using a longer training schedule and with proper regularization and normalization. Li et al. \cite{li2021rethinking} aim to reduce resources and increase stability and propose to split the ``training from scratch'' procedure into a two-step procedure with progressively increasing input resolution, both executed on the target dataset. The combination of training from scratch and data augmentation was explored in \cite{ghiasi2020simple,Xianzhi2021}. Ghiasi et al.~\cite{ghiasi2020simple}, show that the simple data augmentation mechanism of pasting objects randomly onto an image provides solid gains over previous state of the art approaches for both MSCOCO and LVIS detection and segmentation. During their ablation studies, smaller models were trained using strong data augmentation and trained from scratch, while larger models (presenting the best performance) were fine-tuned from a model pre-trained on \ImNet. Du et al.~\cite{Xianzhi2021} present a set of training strategies to improve the detector's performance combining data augmentation (large-scale jittering~\cite{TanPL20}), a smooth activation function, regularization, backbone architecture changes, and normalization techniques. 

\medskip\noindent{\bf Importance of resource savings.} The field of deep learning is becoming increasingly resource-conscious. From an environmental perspective, the carbon footprint of training deep learning models is non-negligible~\cite{thompson2021deep}, and our findings contribute to reducing that footprint. From a social perspective, Obando-Ceron \& Castro~\cite{ceron2021revisiting} note that the large-scale standardized deep reinforcement learning benchmarks such as the Arcade Learning Environment~\cite{bellemare2013arcade} have the unfortunate effect of partitioning the field into groups with access to large-scale computational resources and groups without such resources. They argue that smaller-scale environments can still yield valuable scientific insights while being accessible to more researchers. The large batch sizes used in recent object detection work~\cite{ghiasi2020simple,du2021simple} arguably have a similar deterring effect. While we don't claim to solve the issue in its entirety for object detection, our findings contribute to reducing the computational resources necessary to achieve strong object detection performance and take a step in the direction of allowing researchers with varied levels of access to resources to contribute to the field.

\section{Methodology}

Our main hypothesis is that the features learnt by training on large-scale image classification tasks are better for the object detection task
than those obtained from comparatively smaller, in-domain datasets.  
The classification datasets that we consider (\ImNet (1.2 M) and \JFT (300 M))
contain orders of magnitude more images than common detection datasets like \MSCOCO (118K) and \LVIS (100 K).
They key insight we propose is to freeze the weights learnt on the 
classification task and choose the remaining components such that
they have enough capacity to learn the detection-specific features.

\subsection{Preserving classification features}
To preserve the knowledge learnt during classification, 
we use the most natural and obvious strategy of freezing 
the weights of the classification network (also called backbone).
The common practice in literature \cite{od_survey, fasterrcnn, retinanet} is to
train all the weights in the model after the backbone has
been initialized. We instead consider the alternative
strategy of freezing all of the backbone weights. Not
only does this save compute and speed up training,
but as we discover, improves the performance
of many modern detection architectures.

\subsection{Detection-specific capacity}

Adapting classification networks to the detection task,
typically requires the addition of detection specific components (\autoref{fig:object_detection_diagram}),
like a Region Proposal Network \cite{fasterrcnn}, a Feature
Pyramid Network \cite{fpn} and more recently
Detection Cascades\cite{cascades}.  We observe
that the capacity of detection components plays
a large role in the ability of networks to generalize,
particularly when we initialize from a classification task.
We show that when the detection-specific
components have enough capacity,
initializing from a classification task and freezing those weights performs better than fine-tuning or training from scratch (as is common in the literature). 
Moreover, we see that the performance gain increases when we
pre-train on a more diverse classification dataset.

\subsection{Data augmentation}
We use Large Scale Jittering (LSJ) \cite{TanPL20}
for all of our experiments and Copy-and-paste augmentation
\cite{ghiasi2020simple} for our best results with EfficientNet
\cite{efficientnet}. We note that our proposed
technique is complementary to both these data augmentation
strategies. Moreover, for the experiments with frozen
backbones, the data augmentation techniques are able
to improve results \emph{only} by helping the detection specific
components. 

\section{Experiments}
\label{sec:experiments}

Our experiments connect two apparently contradictory conclusions in the object detection literature.\footnote{See {\tiny\url{https://github.com/tensorflow/models/blob/master/official/projects/backbone_reuse/README.md}} for open-sourced code and instructions.} 
Sun et al. \cite{SunShr17} advocate for pre-training on image classification datasets, and observe benefits that increase with the pretraining dataset's scale. Going in the opposite direction, the most recent trend of papers follows He et al. \cite{He2019RethinkingIP}'s findings that support training from scratch under longer training regimes.

We resolve this contradiction by pointing out that a backbone fine-tuned for longer can move further away from its pre-trained initialization. If---as our evidence suggests---fine-tuning is detrimental to the learned backbone representation in terms of detection performance, then this would explain why the benefits of pre-training appear to vanish in longer training regimes. However, this degradation can be prevented if the pre-trained backbone is frozen during detection training.
This lets us realize the benefits of pre-trained initialization even under longer training schedules.
Doing so is simple, easy to replicate, yields performance improvements on all investigated architectures coupled with expressive-enough subsequent detector components (\autoref{fig:parameter_analysis}), and saves considerable computational resources during training.

\subsection{Experimental Setting}

\noindent{\bf Architecture.} Our baselines were built on top of two open-sourced codebases.
The ablation studies presented using \resnet \cite{HeZRS16} models were built on top of Du et al. \cite{Strategies}'s codebase,\footnote{\tiny\url{https://github.com/tensorflow/models/blob/master/official/vision/beta/MODEL_GARDEN.md}} motivated by the strong performance shown on baselines trained from scratch.
Building on this, our first group of experiments (\autoref{subsec:resnets}) explores the impact of feature preservation on a combinations of small backbones ({\resnet}-50, {\resnet}-101) with {\frcnn}-based detectors. We then introduce detectors with increasing amounts of capacity by varying the feature pyramids \cite{fpn} (\fpn and \nasfpn) and adding \cascade heads \cite{cascades}.

The second group of experiments (\autoref{subsec:effic}) applies the insights gained in a competitive setting, namely a \mrcnn with an \efficientNet backbone and \nasfpn and \cascade heads using Ghiasi et al. \cite{ghiasi2020simple}'s codebase and covers both detection and segmentation tasks.\footnote{\tiny\url{https://github.com/tensorflow/tpu/tree/master/models/official/detection/projects/copy_paste}} We also investigate how Ghiasi et al.'s strong Copy+Paste data augmentation approach interacts with backbone freezing. In comparison to \cite{Strategies}'s baseline,  \cite{ghiasi2020simple}'s implementation replaces {\nasfpn}'s convolution layers with \resnet bottleneck blocks \cite{HeZRS16} and also adds one extra convolutional layers on the RPN head. 

\medskip\noindent{\bf Training Parameters.} All our models trained with frozen backbones adopt exactly the same hyper-parameters as the corresponding non-frozen ones. No changes were made to Ghiasi et al.~\cite{ghiasi2020simple}'s code, while changes to Du et al.~\cite{Strategies}'s code are described in the next paragraph. Further details can be found in the original references.
 
During our study on {\resnet}s models, we target the use and replication of our findings on environments with lower resources available for training. Therefore, the batch size was reduced from Du et al.~\cite{Strategies}'s original value of $256$ down to $64$. Consequently, the learning rate was also adjusted to $0.08$. Results presented in the main text use the same longer training schedule as in their original formulation ($600$ epochs for {\resnet}s and $390$ epochs for {\efficientNet}s). Our results with shorter training schedules are described in \autoref{apd:extra_forshorter}.
 
\medskip\noindent{\bf Datasets.} We perform ablation studies on detection and segmentation with the \MSCOCO (2017)~\cite{lin2014microsoft} and \LVIS $1.0$~\cite{lvis} datasets. \MSCOCO has $118k$ images covering $91$ classes ($80$ used in practice).
 The \LVIS dataset was designed to simulate the long-tail distribution of classes in natural images. The $1.0$ version used in this paper contains $100k$ images covering $1203$ classes. 

Due to large costs associated with annotating localization information, image classification datasets are much larger than object detection \cite{Papadopoulos2017ExtremeCF}
and segmentation \cite{lin2014microsoft} datasets. Since we investigate the benefits associated with dataset scale, we use two image classification datasets with increasing amounts of data.  
 \ImNet contains $1M$ labeled images based on $1000$ categories while \JFT \cite{Hinton15} contains more than $300M$ images labeled with $18291$ categories. Labeling error in \JFT is estimated to be around $20\%$~\cite{SunShr17}, but this is offset by a larger visual diversity than \ImNet.
 We use \ImNet or \JFT to pretrain the backbone on classification tasks,
 except when training from scratch.
 Neither is used to train the detector.

\subsection{Revisiting {\resnet} reuse for object detection}
\label{subsec:resnets}

\begin{table}
\centering
\footnotesize
\begin{tabular}{llrr}
  \toprule
    {\bf Model} & {\bf Pretraining} & \multicolumn{1}{r}{\bf mAP} & \multicolumn{1}{r}{\bf  AP $@$ 50} \\ 
  \midrule
    \multirow{5}{*}{\makecell{ResNet-101 \\+ FPN}} & \Scratch  &  $48.4 $  & $70.1$\\
     & \Imagenet  &  $48.6 $ & $70.5 $ \\
     & \hspace{0.5em} + Freeze backbone & \mathdeltaneg{-6.5} $42.1 $ & \mathdeltaneg{-5.2} $65.3$\\
     & \JFT  &  $48.7 $  & $70.5 $\\
     & \hspace{0.5em} + Freeze backbone & \mathdeltaneg{-5.6  } $43.1 $ & \mathdeltaneg{-3.3} $67.2$\\
  \midrule
    \multirow{5}{*}{\makecell{ResNet-101\\+ \nasfpn}} & \Scratch  &  $47.2 $  & $68.2 $\\ 
     & \Imagenet  &  $49.0 $ & $70.0 $ \\
     & \hspace{0.5em} + Freeze backbone & \mathdelta{+0.1} $49.1  $ & \mathdelta{+0.3} $70.3$\\
     & \JFT  &  $49.1 $  & $70.2$\\
     & \hspace{0.5em} + Freeze backbone & \mathdelta{+1.0} $50.1$ & \mathdelta{+1.5} $ 71.8$\\
  \midrule
    \multirow{5}{*}{\makecell{ResNet-50 \\+ \nasfpn \\+ Cascade}} & \Scratch  &  $50.0 $  & $68.1$\\
     & \Imagenet  &  $50.3 $ & $68.0 $ \\
     & \hspace{0.5em} + Freeze backbone & \mathdelta{+0.7} $51.0 $ & \mathdelta{+1.1} $69.1$\\
     & \JFT  &  $50.4 $  & $68.1 $\\
     & \hspace{0.5em} + Freeze backbone & \mathdelta{+1.7} $52.1 $ & \mathdelta{+2.3} $70.4 $\\
  \midrule
    \multirow{5}{*}{\makecell{ ResNet-101 \\+ \nasfpn \\+ Cascade}} & \Scratch  &  $50.4 $  & $68.4 $\\
     & \Imagenet  &  $51.1$ & $69.1  $ \\
     & \hspace{0.5em} + Freeze backbone & \mathdelta{+0.8} $ 51.8 $ & \mathdelta{0.8} $69.9 $\\
     & \JFT  &  $51.1 $  & $69.0 $\\
     & \hspace{0.5em} + Freeze backbone & \mathdelta{+1.7 } $52.8$ & \mathdelta{+2.1} $71.1 $\\
  \bottomrule
\end{tabular}
\caption{\label{tab:resnet_longer} With a powerful-enough detector, freezing the backbone to its pre-trained initialization during detection training outperforms fine-tuning the backbone or training it from scratch.}
\end{table}

We start by investigating the impact of the backbone training strategy (trained from scratch, fine-tuned from a pre-trained initialization, or frozen at its pre-trained initialization) while controlling for pre-training dataset size (\Imagenet, \JFT), backbone architecture (\resnet-50, \resnet-101), detector architecture (\fpn, \nasfpn, \nasfpn + \cascade), and training schedule (72 epochs, 600 epochs). Tabular results for the longer training schedule are presented and discussed here (\autoref{tab:resnet_longer}), and results using the shorter schedule are presented in \autoref{apd:extra_forshorter}.


\begin{table}
\centering
\footnotesize
\begin{tabular}{lrrrr}
  \toprule
{\bf Paper} & {\bf Pre-training} & {\bf Schedule} & {\bf Freeze?} & {\bf mAP} \\
\midrule
\multicolumn{5}{c}{\em Pre-training on larger classification datasets helps.\vspace{0.5em}} \\
\multirow{2}{*}{Sun et al.~\cite{SunShr17}} & ImageNet & Short & Yes & 47.8 \\
& JFT & Short & Yes & 49.0 \\
\midrule
\multicolumn{5}{c}{\em Pre-training does not help with longer training schedules.\vspace{0.5em}} \\
\multirow{2}{*}{He et al.~\cite{He2019RethinkingIP}} & ImageNet & Long & No & 49.0 \\
& JFT & Long & No & 49.1 \\
\midrule
\multicolumn{5}{c}{\em Backbone freezing \& high-capacity detector components help.\vspace{0.5em}} \\
Ours & JFT & Long & Yes & {\bf 50.1} \\
\bottomrule
\end{tabular}
\caption{\label{fig:longer_vs_shorter}
We revisit the conclusions from~\cite{He2019RethinkingIP,SunShr17} under modern training regimes and best practices. We use their main conclusions and re-train these models for a fair comparison. All results are reported on a ResNet-101 backbone with a NAS-FPN. Note that Sun et al.~\cite{SunShr17} did not run experiments with NAS-FPN. }
\end{table}

\medskip\noindent{\bf The relative benefit from pre-training on larger classification datasets is clear when freezing the backbone.} The comparison between {\em \Imagenet + Freeze backbone} and  {\em \JFT + Freeze backbone} in \autoref{tab:resnet_longer} shows a consistent improvement in performance ($+0.9$ mAP in most cases) across the different backbones and detector components tested. See \autoref{apd:capacity} (\autoref{fig:relative_improvement}) for an alternative visualization.

\medskip\noindent{\bf Proper reuse of image classification features improves performance.}
Our main finding is that the advantage gained from pre-trained backbone networks gets lost with a longer training schedule. Notice that in \autoref{fig:longer_vs_shorter}, the improvement from \JFT which was apparent in the short training schedule (blue-shaded region), disappears with the long training schedule (red-shaded region). In contrast, we show that with  freezing the backbone, and thus preserving the pre-training knowledge, we  maintain the improvement due to \JFT across both the training schedules.

Our results corroborate Sun et al.~\cite{SunShr17}'s observation that pre-training on larger datasets is beneficial when fine-tuning the backbone for a shorter
schedule (\autoref{fig:longer_vs_shorter}, shadowed blue), and \autoref{apd:extra_forshorter}'s \autoref{tab:resnet_shorter} shows this is consistent across architectures.
Our results complement Sun et al.~\cite{SunShr17} and He et al.~\cite{He2019RethinkingIP}: by freezing the backbone, we see that both can be explained through the lens of pre-trained knowledge preservation.
In addition, we notice that the benefit from the re-use of the knowledge from large scale image classification datasets is bounded by the capacity of remaining components, as we show next.

\begin{figure}
    \centering
    \includegraphics[width=\linewidth]{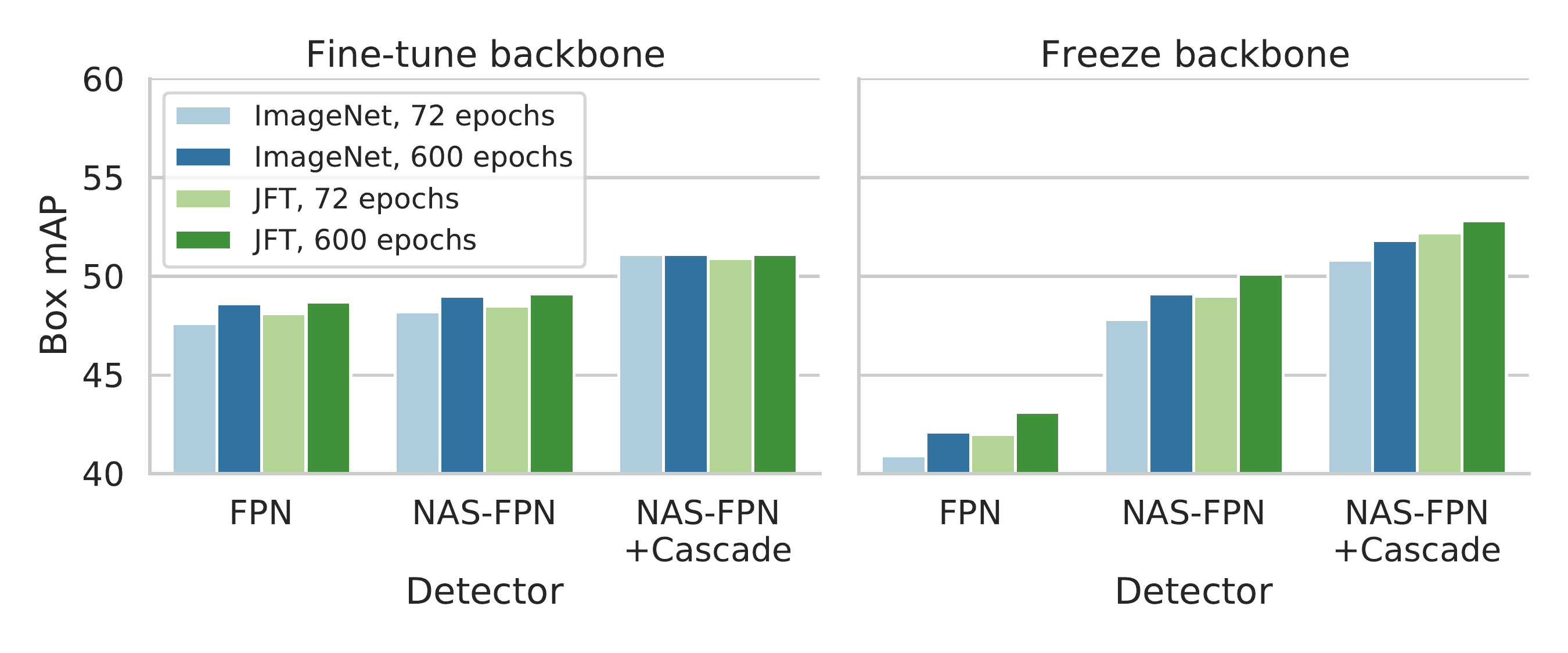}
    \caption{\label{fig:imagenet_vs_jft} Detectors using a {\resnet}-101 backbone. Sufficient capacity on components trained is important in realizing the benefits of preserving the representation from a large classification dataset.}
\end{figure}

\medskip\noindent{\bf Feature preservation benefits are bounded by the remaining trainable  capacity.} 
Sun et al.~\cite{SunShr17} conjecture that the overall benefit from pre-training is bounded by the capacity of the whole model (backbone + detection components). Using an \fpn detector, they present experiments freezing backbone weights under short training regimes and show that those models under-perform their fine-tuned counterparts. 
Our results using \fpn detectors confirm the same decrease in performance 
(\autoref{fig:imagenet_vs_jft}) on both short and long training schedules.

In a new direction of investigation, we refine the original conjecture to pinpoint the remaining detector components as the principal capacity bottleneck in terms of benefiting from pre-trained features. \autoref{fig:parameter_analysis} plots the performance observed across models built with components of increasing capacity and trained with either backbone freezing or fine-tuning. \autoref{tab:num_params} shows the number of parameters of the models ablated and the corresponding number of parameters trained on models with frozen backbones. 
Object detection models with higher capacity in the remaining trainable components clearly surpass their fine-tuned counterparts. 
See \autoref{apd:capacity} for more capacity ablations.

To summarize, 
in typical settings (e.g. using an FPN) we don't see benefits from freezing the backbone
because the remaining detector components don't have enough capacity.
In other words, the representation learned from pre-training on the large
classification dataset is better, provided that there are enough learnable
parameters (like with NAS-FPN).  
Consequently, the comparable performances achieved by fine-tuning the backbone and training it from scratch under a long schedule could simply be a consequence of the fact that fine-tuning for longer moves it further away from the good representation found by pre-training on the classification task.
 
Through backbone freezing and extensive experimentation, we 
 (i) disentangle the benefit of using pre-trained backbones when compared to training from scratch under long training schedules; and (ii) show the benefit on the final detector of using even larger classification datasets.
Additionally, our results suggest that the knowledge contained in pre-trained weights can and should be preserved during longer training regimes.

\begin{table}
\centering
\footnotesize
\label{tab:params}
\begin{tabular}{lrrr}
  \toprule
{\bf Detection model} & \multicolumn{2}{c}{\bf \#Params (Million)} & $\mathbf{\Delta}$ {\bf mAP} \\
\cmidrule{2-3}
& {\bf Original} & {\bf Trained} \\
   \midrule
\fpn & $83.5$ &  ($26.1\%$) $21.9$ & $-6.5$\\
\nasfpn & $102.6$ &  $(39.9\%$) $40.9$ & $+0.1$\\
Cascade + \nasfpn  & $132.9$ &  $(53.6\%$) $71.3$ & $+0.7$\\
\bottomrule
  
\end{tabular}
\caption{\label{tab:num_params}
Number of parameters that are trained when keeping the backbone frozen (a \resnet-101) with various detection models. We report  $\mathbf{\Delta}$ mAP as the performance gap from fine-tuned counterparts. Weights initialized from an
\ImNet pre-trained model.}
\end{table}
\subsection{{\efficientNet}'s reuse for object detection}
\label{subsec:effic}

Next, we present our results on freezing features learned from a classification  dataset (\Imagenet) on \efficientNet based detectors. 
The ablations also investigate how freezing the backbone complements strong data augmentation techniques designed specifically for localization tasks. The models compared in this section adopt both large scale jittering \cite{TanPL20} and Copy+Paste \cite{ghiasi2020simple} data augmentation (previously shown to improve training under fine-tuning and training from scratch regimes). 
The comparison of our results with other strong  detectors are presented in \autoref{tab:sota}. 

\medskip\noindent{\bf Freezing \efficientNet backbones improves performance on both  \MSCOCO and \LVIS datasets}. 
\autoref{tab:sota} presents our results when applied to the same architectures and training regimes as Ghiasi et al.~\cite{ghiasi2020simple}'s baselines for \MSCOCO and \LVIS. 
It shows a clear gain in performance for frozen models observed both on detection and
instance segmentation tasks. 
On the \MSCOCO baseline, backbone freezing outperforms fine-tuning by $+0.9$  Box mAP. Note that this number on the impact of freezing alone is comparable to the benefit of using Copy+Paste augmentation ($+1.1$). 
From the same table, the results on \MSCOCO instance segmentation highlight the benefit of feature preservation even more, with a larger benefit derived ($+0.7$ Mask mAP) than from Copy+Paste ($+0.3$).

Similarly, freezing classification features also presented a positive impact on performance for \LVIS for both detection and instance segmentation. This strengthens our claim that feature preservation is important, since \LVIS is harder than \MSCOCO due to its long-tailed classed distribution.

\medskip\noindent{\bf When the backbone is frozen, strong data augmentations improves the remaining detector components even further.}
The benefit of feature preservation is even more clear when combined
with localization-specific data augmentations. 
For \MSCOCO detection, the improvement obtained by feature preservation is of $+1.1$ mAP and reaches $+1.5$ mAP for segmentation masks (\autoref{tab:sota}). 
This increase in segmentation performance is larger than the original $+0.3$ gain obtained by using the Copy+Paste augmentation itself ($+1.1$ Box mAP and $+0.3$ Mask mAP).
That is, by combining  feature preservation and Copy+Paste augmentation the total gain over the baseline numbers is $+2.2$ Box mAP and $+1.8$ Mask AP. 
Note that in this case the data augmentations only affect the detection-specific
components, since the backbone remains frozen.

\begin{table}
\centering
\footnotesize
\begin{tabular}{lrr}
  \toprule
    {\bf MSCOCO} (val) & {\bf Box mAP} & {\bf Mask mAP} \\
  \midrule
    Swin Transformer~\cite{liu2021swin}$^*$ & {\bf 57.1} & {\bf 49.5} \\
    Cascade Eff-B7 \nasfpn (1280)~\cite{ghiasi2020simple} & 54.8 & 46.9 \\
    \hspace{0.5em} {\bf + freeze backbone (ours)} &  \mathdelta{+0.9} 55.7 & \mathdelta{+0.7} 47.6 \\
    \hspace{0.5em} + Copy-Paste & 55.9 & 47.2 \\
    \hspace{1em} {\bf + freeze backbone (ours)} & \mathdelta{+1.1} 57.0 & \mathdelta{+1.5} 48.7 \\
  \midrule
    \hspace{0.5em} + Self-training Copy-Paste & 57.0 & 48.9 \\
    Soft Teacher + Swin-L~\cite{xu2021end} & 59.1 & 51.0 \\
  \bottomrule
     & & \\
  \toprule
    {\bf LVIS} (val) & {\bf Box mAP} & {\bf Mask mAP} \\
  \midrule
    cRT (ResNeXt-101-32$\times$8d)~\cite{kang2019decoupling} & --- & 27.2 \\
    LVIS Challenge 2020 Winner~\cite{tan20201st}$^\dagger$ & 41.1 & 38.8 \\
    Eff-B7 \nasfpn (1280)~\cite{ghiasi2020simple} & 37.2 & 34.7 \\
    \hspace{1em} {\bf + freeze backbone (ours)} &  \mathdelta{+2.2} 39.4 &  \mathdelta{+2.5} 37.2  \\ 
    \hspace{0.5em} + Copy-Paste & 41.6 & 38.1 \\
    \hspace{1em} {\bf + freeze backbone (ours)} & {\bf \mathdelta{+1.5} 43.1} & {\bf \mathdelta{+1.8} 39.9}  \\
  \bottomrule
\end{tabular}
\caption{\label{tab:sota}
Freezing the backbone
improves two strong baselines on \MSCOCO and \LVIS.
when training only on the \texttt{train2017} split.
``Self-training'' and \cite{xu2021end}~use the extra training data. $^*$ImageNet-22K pre-training. $^\dagger$No test-time augmentation.
All results are reported on the validation sets.
}
\end{table}

The combination of feature preservation and data augmentation produces even larger improvements in performance on \LVIS (\autoref{tab:sota}). This can be observed on both detection ($+1.5$ over Copy-Paste and $+5.9$ over the baseline) and segmentation 
($+1.8$ over Copy-Paste and $+5.2$ over the baseline). 
We further stratify the results obtained on \LVIS by number of annotations (\autoref{tab:lvis_num_annotations}) and object size (\autoref{tab:lvis_object_size}). Our \efficientNet models with a frozen pre-trained backbone show improvements across all object sizes, numbers of annotations, and detection and segmentation tasks.

Our \LVIS results are obtained by following Ghiasi et al.~\cite{ghiasi2020simple}'s two-step training procedure: after training the detector using unbalanced loss, the classifier head is further tuned using a class balanced loss. The goal of the second stage is to improve performance on rare classes. 
In the original fine-tuning setting the second training stage causes a drop in performance on the \emph{frequent} classes.
However, with feature preservation, all three groups (rare, common and frequent classes)
improve in performance in the second stage.


\begin{table*}
\centering
\footnotesize
\begin{tabular}{lrrrrrrrr}
  \toprule
   LVIS  & \multicolumn{4}{c}{\bf Box} & \multicolumn{4}{c}{\bf Mask} \\
  \cmidrule(r){2-5} \cmidrule(l){6-9} 
    & {\bf mAP$_{\mathrm{\bf}}$}  & {\bf mAP$_{\mathrm{\bf r}}$} & {\bf mAP$_{\mathrm{\bf c}}$} & {\bf mAP$_{\mathrm{\bf f}}$} & {\bf mAP$_{\mathrm{\bf}}$}  & {\bf mAP$_{\mathrm{\bf r}}$} & {\bf mAP$_{\mathrm{\bf c}}$} & {\bf mAP$_{\mathrm{\bf f}}$}  \\
  \midrule
    \multicolumn{9}{l}{\bf First stage results: regular training} \\
    Copy and Paste \cite{ghiasi2020simple} & $38.5$ & $19.3$ & $37.3$ & $48.2$ & $35.0$ & $19.5$ & $34.9$ & $42.1$ \\
    \hspace{0.5em} + Freeze backbone & \mathdelta{+1.2} $39.7$ & \mathdelta{+2.9} $22.2$ & \mathdelta{+1.5} $38.8$ & \mathdelta{+0.1} $48.3$ & \mathdelta{+1.5} $36.5$ & \mathdelta{+2.9} $22.4$ & \mathdelta{+1.9} $36.8$ & \mathdelta{+0.4} $42.5$\\
  \midrule
  \multicolumn{9}{l}{\bf Second stage: tunes detection-classifier final layer using class-balanced loss} \\
    Copy and Paste \cite{ghiasi2020simple} & $41.6$ & $31.5$ & $39.8$ & $48.0$ & $38.1$ & $32.1$ & $37.1$ & $41.9$ \\
    \hspace{0.5em} + Freeze backbone & \mathdelta{+1.5} $43.1$ & \mathdelta{+1.7} $33.2$ & \mathdelta{+2.1} $41.9$ & \mathdelta{+0.7} $48.7$ &
    \mathdelta{+1.1} $39.9$ & \mathdelta{+1.5} $33.6$ & \mathdelta{+2.5} $39.6$ & \mathdelta{+1.0} $42.9$ \\
  \bottomrule
\end{tabular}
\caption{\label{tab:lvis_num_annotations} Performance using \efficientNet + \nasfpn.
Freezing the backbone has the strongest positive performance impact on rare ($\mathrm{mAP}_\mathrm{r}$) and common ($\mathrm{mAP}_\mathrm{c}$) classes, while still improving frequent ($\mathrm{mAP}_\mathrm{f}$) classes. Original first phase results are provided by the authors of \cite{ghiasi2020simple}. Results without Copy-Paste augmentations can be found in \autoref{apd:extra_lvis}.}
\end{table*}

\begin{table*}
\centering
\footnotesize
\begin{tabular}{lrrrrrrrr}
  \toprule
  LVIS   & \multicolumn{4}{c}{\bf Box} & \multicolumn{4}{c}{\bf Mask} \\
  \cmidrule(r){2-5} \cmidrule(l){6-9} 
    & {\bf mAP$_{\mathrm{\bf}}$}  & {\bf mAP$_{\mathrm{\bf s}}$} & {\bf mAP$_{\mathrm{\bf m}}$} & {\bf mAP$_{\mathrm{\bf l}}$} & {\bf mAP$_{\mathrm{\bf}}$}  & {\bf mAP$_{\mathrm{\bf s}}$} & {\bf mAP$_{\mathrm{\bf m}}$} & {\bf mAP$_{\mathrm{\bf l}}$}  \\
  \midrule
    \multicolumn{9}{l}{\bf First stage results: regular training} \\
    Copy and Paste \cite{ghiasi2020simple} &  $ 38.5$ &$30.5$ &$ 48.2$ &$55.5$ & $ 35.0$ &$ 25.7$ &$ 45.6$ &$ 53.2$ \\
    \hspace{0.5em} + frozen backbone & \mathdelta{+1.2} $39.7$ & \mathdelta{+0.2} $30.7$& \mathdelta{+1.4} $49.6$& \mathdelta{+2.3} $57.8$ &   \mathdelta{+1.5} $36.5$ & \mathdelta{+0.3} $ 26.0$ & \mathdelta{+1.6} $ 47.2$ & \mathdelta{+2.8} $ 56.0$\\
  \midrule
  \multicolumn{9}{l}{\bf Second stage: tunes detection-classifier final layer using class-balanced loss} \\
    Copy and Paste \cite{ghiasi2020simple} &  $41.6$ &$33.5$ &$51.5$ &$58.1$ & $ 38.1$ &$ 28.4$ &$ 49.0$ &$ 55.6$ \\
    \hspace{0.5em} + frozen backbone & \mathdelta{+1.5} $43.1$ & \mathdelta{+0.6}$34.1$ & \mathdelta{+1.8} $53.3	$& \mathdelta{+2.3} $60.4$ &  \mathdelta{+1.8} $ 39.9$ & \mathdelta{+0.7} $ 29.1$ & \mathdelta{+2.1} $51.1$ & \mathdelta{+2.7} $	58.3$\\
  \bottomrule
\end{tabular}
\caption{\label{tab:lvis_object_size} Performance using \efficientNet + \nasfpn. Freezing the backbone has the strongest positive performance impact on large objects ($\mathrm{mAP}_\mathrm{l}$), then medium-sized objects ($\mathrm{mAP}_\mathrm{m}$), and finally small objects ($\mathrm{mAP}_\mathrm{s}$). Original first phase results are provided by the authors of \cite{ghiasi2020simple}. Results without Copy-Paste augmentations can be found in \autoref{apd:extra_lvis}.}
\end{table*}

\subsection{How does preserving pre-trained representations help?}

\begin{figure}
    \centering
    \includegraphics[width=\linewidth]{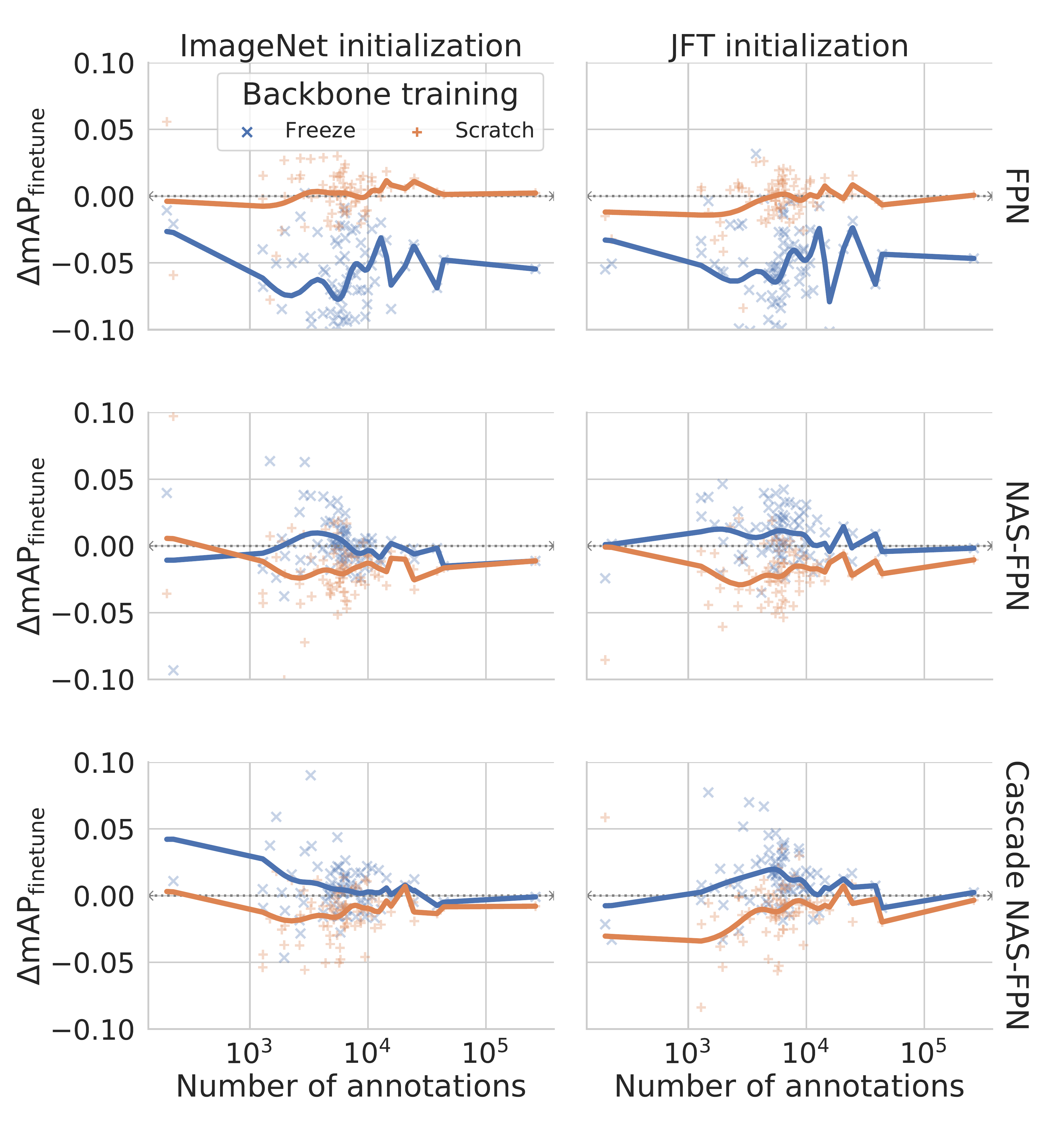}
    \caption{\label{fig:class_wise_analysis_coco}Class-wise mAP as a function of the number of training annotations {\em relative to fine-tuning the backbone}. Smoothed curves are obtained by applying Gaussian smoothing with $\sigma = 1000$. The ResNet-101 backbone is combined with detectors of varying  capacities (rows) and initializations (columns). Freezing the backbone is increasingly beneficial as the detector capacity increases (top to bottom rows) and as the number of annotations decreases.}
\end{figure}

\begin{figure}
    \centering
    \includegraphics[width=\linewidth]{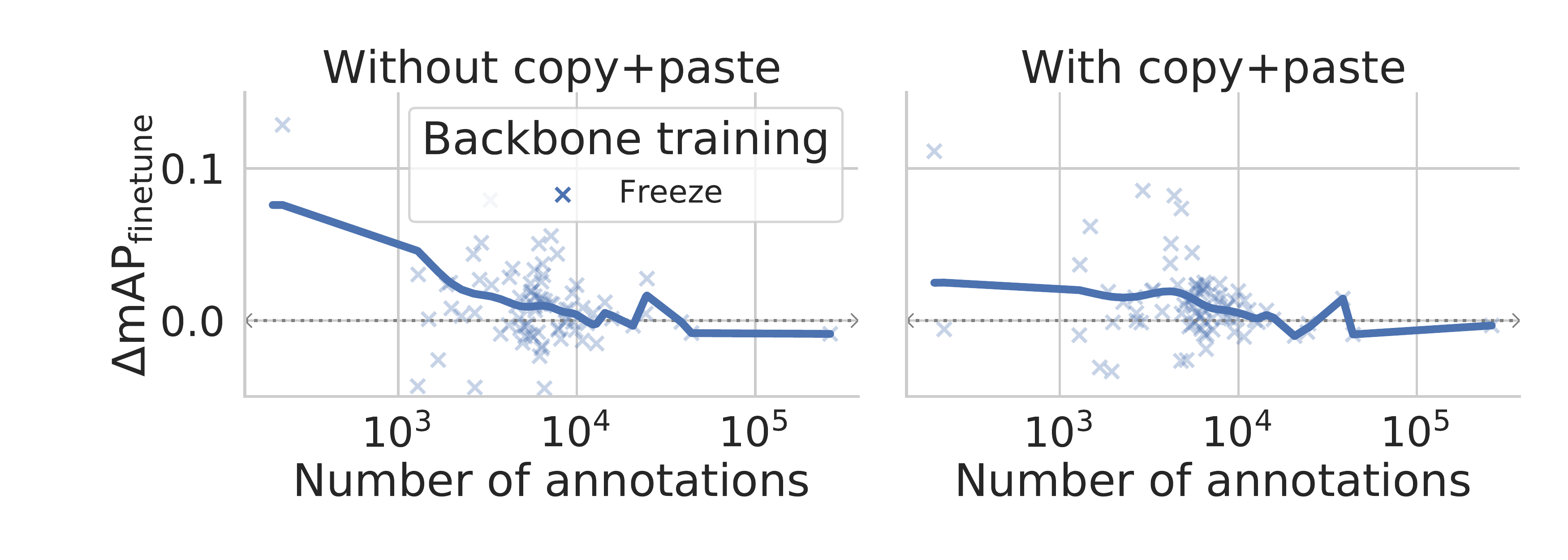}
    \caption{\label{fig:class_wise_analysis_coco_copy_paste}Class-wise mAP as a function of the number of training annotations {\em relative to fine-tuning the backbone}. Smoothed curves are obtained by applying Gaussian smoothing with $\sigma = 1000$. The \efficientNet architecture is trained without (left) and with (right) copy+paste augmentations.}
\end{figure}

So far we have established that given sufficient detector capacity it is better to freeze the pre-trained backbone than to fine-tune it or train it from scratch on detection data, but it is not obvious why that is the case.

We explore this question by visualizing how freezing the backbone impacts performance on classes with different numbers of training annotations. \autoref{fig:class_wise_analysis_coco} shows class-wise box mAPs relative to those obtained by fine-tuning the backbone for the {\em freeze} and {\em from scratch} training strategies. As before, we observe that with a lower-capacity detector (\fpn) training the backbone from scratch is comparable in performance to fine-tuning it from a pre-trained initialization, and that freezing the backbone underperforms fine-tuning it. Interestingly, fine-tuning from a JFT initialization outperforms training from scratch for classes with fewer annotations. As we move towards larger detectors, performance across the three strategies remains similar for classes with larger amounts of annotations while freezing the backbone and training it from scratch become increasingly beneficial and detrimental (respectively) for classes with fewer annotations.

We observe a similar behaviour when comparing {\em freeze} and {\em fine-tune} in a more competitive setting (\autoref{fig:class_wise_analysis_coco_copy_paste}, \autoref{tab:lvis_num_annotations}), where the benefits of backbone freezing concentrate mostly on classes with fewer annotations.

Given these observations, we conjecture that the pre-trained representation contains features beneficial for detection that require many annotations to be learned from the detection data alone in addition to being brittle to fine-tuning on detection data. While not identical to catastrophic forgetting~\cite{FRENCH1999128}, this phenomenon bears some resemblance to it: in fine-tuning on the detection task, the object detection model appears to struggle to preserve knowledge not only beneficial to the upstream classification task, but {\em beneficial to the downstream detection task itself}.

\subsection{Beyond backbone freezing}
\label{subsec:adapters}

Throughout this work we presented backbone freezing as a knowledge preservation strategy that yields performance benefits for object detection. What this shows is that there are better ways of using a pre-trained model for downstream object detection applications, but it does not mean (nor do we claim) that backbone freezing is itself an optimal strategy.

To demonstrate this, we present initial results using a lightweight alternative strategy in the form of residual adapters~\cite{RebuffiBV17,Rebuffi18}, which have been successfully applied to adapt to downstream tasks such as cross-domain few-shot image classification~\cite{li2021improving}, natural language processing~\cite{Houlsby19,pfeiffer2021adapterfusion,Mahabadi21}, and transfer learning with expert image classifiers~\cite{Puigcerver2021ScalableTL}.

Our initial results (\autoref{tab:resnet_residual}) show that equipping the frozen backbone with residual adapter layers that are trained alongside the detector components yields further improvements over full backbone fine-tuning. We conjecture that residual adapters and similar adaptation approaches such as feature-wise transformations~\cite{dumoulin2018feature} can better incorporate the object detection training signal while preserving aspects of the image classifier representation that are useful to the object detection task. See \autoref{apd:adapters} for more results on the use of adapters.

More generally, adjacent fields such as transfer learning, multi-task learning, few-shot classification, and domain adaptation have all tackled the problem of using pre-trained models for downstream applications, and we believe that object detection would benefit from their insights.

\begin{table}
\centering
\footnotesize
\begin{tabular}{llrr}
  \toprule
    {\bf Model} & {\bf Pretraining} & \multicolumn{1}{r}{\bf mAP} & \multicolumn{1}{r}{\bf  AP $@$ 50} \\
  \midrule
     \multirow{6}{*}{\makecell{ResNet-101 \\+ \nasfpn \\+ Cascade}} & \Imagenet  &  $51.1$ & $69.1  $ \\
     & \hspace{0.5em} + Freeze backbone & \mathdelta{+0.8} $ 51.8 $ & \mathdelta{+0.8} $69.9 $\\
     & \hspace{1.0em} + Res. adapters & \mathdelta{+1.9} $ 53.0 $ & \mathdelta{+2.4} $71.5$\\
     & \JFT  &  $51.1 $  & $69.0 $\\
     & \hspace{0.5em} + Freeze backbone & \mathdelta{+1.7} $52.8$ & \mathdelta{+2.1} $71.1 $\\
     & \hspace{1.0em} + Res. adapters & \mathdelta{+2.5} $53.6$ & \mathdelta{+3.0} $72.0 $\\
  \bottomrule
\end{tabular}
\caption{\label{tab:resnet_residual} Adding residual adapters to the frozen backbone and training them along with the subsequent detector components yields another performance increase.}
\end{table}
\section{Conclusions}
\label{sec:conclusion}

We cast a new light on the re-use of pre-trained representations obtained
from a large-scale classification task in a downstream detection setting.
Particularly, we show that preserving the backbone representation obtained
from training on a large-scale classification task is beneficial to object
detection and instance segmentation. 
We also show how this ties together the two seemingly contradictory observations
of \cite{SunShr17} and \cite{He2019RethinkingIP}, the missing piece being the fact that the longer training schedule moves the backbone further away from a good initial representation.

We investigate backbone freezing as a simple approach to knowledge preservation and demonstrate its benefits when coupled with sufficient detection-specific component capacity through extensive experiments across multiple combinations of backbones, detection models, datasets, and training schedules. This approach easy to implement and reproduce and requires significantly less computational resources during training.

While the need to control for resources used in previous work means that SOTA is not yet accessible to most,  we demonstrate demonstrate computation and memory savings in all settings, meaning that practitioners are able to train larger models with larger batch sizes given the same amount of resources, and accessibility could be further improved through future work on reducing the number of training epochs required. We also believe that tapping into the rich model re-use literature in adjacent fields represents a promising direction for future work. Finally, our findings can be used in future neural architecture search work to take advantage of pre-trained and frozen classification based features to ultimately do better than \nasfpn. 

\pagebreak


{\small
\bibliographystyle{ieee_fullname}
\bibliography{adapters}

\begin{thebibliography}{10}\itemsep=-1pt

\bibitem{bar2021detreg}
Amir Bar, Xin Wang, Vadim Kantorov, Colorado~J Reed, Roei Herzig, Gal Chechik,
  Anna Rohrbach, Trevor Darrell, and Amir Globerson.
\newblock Detreg: Unsupervised pretraining with region priors for object
  detection, 2021.

\bibitem{bellemare2013arcade}
Marc~G Bellemare, Yavar Naddaf, Joel Veness, and Michael Bowling.
\newblock The arcade learning environment: An evaluation platform for general
  agents.
\newblock {\em Journal of Artificial Intelligence Research}, 47:253--279, 2013.

\bibitem{cascades}
Zhaowei Cai and Nuno Vasconcelos.
\newblock Cascade r-cnn: Delving into high quality object detection.
\newblock In {\em Proceedings of the IEEE conference on computer vision and
  pattern recognition}, pages 6154--6162, 2018.

\bibitem{ceron2021revisiting}
Johan Samir~Obando Ceron and Pablo~Samuel Castro.
\newblock Revisiting rainbow: Promoting more insightful and inclusive deep
  reinforcement learning research.
\newblock In {\em International Conference on Machine Learning}, pages
  1373--1383. PMLR, 2021.

\bibitem{chollet2017xception}
Fran{\c{c}}ois Chollet.
\newblock Xception: Deep learning with depthwise separable convolutions.
\newblock In {\em Proceedings of the IEEE conference on computer vision and
  pattern recognition}, pages 1251--1258, 2017.

\bibitem{Dai_2021_CVPR}
Zhigang Dai, Bolun Cai, Yugeng Lin, and Junying Chen.
\newblock Up-detr: Unsupervised pre-training for object detection with
  transformers.
\newblock In {\em Proceedings of the IEEE/CVF Conference on Computer Vision and
  Pattern Recognition (CVPR)}, pages 1601--1610, June 2021.

\bibitem{Xianzhi2021}
Xianzhi Du, Barret Zoph, Wei{-}Chih Hung, and Tsung{-}Yi Lin.
\newblock Simple training strategies and model scaling for object detection.
\newblock {\em CoRR}, abs/2107.00057, 2021.

\bibitem{du2021simple}
Xianzhi Du, Barret Zoph, Wei-Chih Hung, and Tsung-Yi Lin.
\newblock Simple training strategies and model scaling for object detection,
  2021.

\bibitem{Strategies}
Xianzhi Du, Barret Zoph, Wei-Chih Hung, and Tsung-Yi Lin.
\newblock Simple training strategies and model scaling for object detection.
\newblock {\em arXiv preprint arXiv:2107.00057}, 2021.

\bibitem{dumoulin2018feature}
Vincent Dumoulin, Ethan Perez, Nathan Schucher, Florian Strub, Harm~de Vries,
  Aaron Courville, and Yoshua Bengio.
\newblock Feature-wise transformations.
\newblock {\em Distill}, 3(7):e11, 2018.

\bibitem{FRENCH1999128}
Robert~M. French.
\newblock Catastrophic forgetting in connectionist networks.
\newblock {\em Trends in Cognitive Sciences}, 3(4):128--135, 1999.

\bibitem{ghiasi2020simple}
Golnaz Ghiasi, Yin Cui, Aravind Srinivas, Rui Qian, Tsung-Yi Lin, Ekin~D Cubuk,
  Quoc~V Le, and Barret Zoph.
\newblock Simple copy-paste is a strong data augmentation method for instance
  segmentation.
\newblock {\em arXiv preprint arXiv:2012.07177}, 2020.

\bibitem{lvis}
Agrim Gupta, Piotr Doll{\'{a}}r, and Ross~B. Girshick.
\newblock {LVIS:} {A} dataset for large vocabulary instance segmentation.
\newblock In {\em {IEEE} Conference on Computer Vision and Pattern Recognition,
  {CVPR} 2019, Long Beach, CA, USA, June 16-20, 2019}, pages 5356--5364.
  Computer Vision Foundation / {IEEE}, 2019.

\bibitem{He2019RethinkingIP}
Kaiming He, Ross Girshick, and Piotr Doll{\'a}r.
\newblock Rethinking imagenet pre-training.
\newblock In {\em Proceedings of the International Conference on Computer
  Vision}, pages 4918--4927, 2019.

\bibitem{HeZRS16}
Kaiming He, Xiangyu Zhang, Shaoqing Ren, and Jian Sun.
\newblock Deep residual learning for image recognition.
\newblock In {\em 2016 {IEEE} Conference on Computer Vision and Pattern
  Recognition, {CVPR} 2016, Las Vegas, NV, USA, June 27-30, 2016}, pages
  770--778. {IEEE} Computer Society, 2016.

\bibitem{hinton2015distilling}
Geoffrey Hinton, Oriol Vinyals, and Jeff Dean.
\newblock Distilling the knowledge in a neural network.
\newblock {\em arXiv preprint arXiv:1503.02531}, 2015.

\bibitem{Hinton15}
Geoffrey Hinton, Oriol Vinyals, and Jeffrey Dean.
\newblock Distilling the knowledge in a neural network.
\newblock In {\em NIPS Deep Learning and Representation Learning Workshop},
  2015.

\bibitem{Houlsby19}
Neil Houlsby, Andrei Giurgiu, Stanislaw Jastrzebski, Bruna Morrone, Quentin de
  Laroussilhe, Andrea Gesmundo, Mona Attariyan, and Sylvain Gelly.
\newblock Parameter-efficient transfer learning for {NLP}.
\newblock {\em CoRR}, abs/1902.00751, 2019.

\bibitem{Huang16}
Jonathan Huang, Vivek Rathod, Chen Sun, Menglong Zhu, Anoop Korattikara,
  Alireza Fathi, Ian Fischer, Zbigniew Wojna, Yang Song, Sergio Guadarrama, and
  Kevin Murphy.
\newblock Speed/accuracy trade-offs for modern convolutional object detectors.
\newblock {\em CoRR}, abs/1611.10012, 2016.

\bibitem{od_survey}
Jonathan Huang, Vivek Rathod, Chen Sun, Menglong Zhu, Anoop Korattikara,
  Alireza Fathi, Ian Fischer, Zbigniew Wojna, Yang Song, Sergio Guadarrama, and
  Kevin Murphy.
\newblock Speed/accuracy trade-offs for modern convolutional object detectors.
\newblock In {\em Proceedings of the IEEE Conference on Computer Vision and
  Pattern Recognition (CVPR)}, July 2017.

\bibitem{Jeong_2019}
Jisoo Jeong, Seungeui Lee, Jeesoo Kim, and Nojun Kwak.
\newblock Consistency-based semi-supervised learning for object detection.
\newblock In H. Wallach, H. Larochelle, A. Beygelzimer, F. d\textquotesingle
  Alch\'{e}-Buc, E. Fox, and R. Garnett, editors, {\em Advances in Neural
  Information Processing Systems}, volume~32. Curran Associates, Inc., 2019.

\bibitem{kang2019decoupling}
Bingyi Kang, Saining Xie, Marcus Rohrbach, Zhicheng Yan, Albert Gordo, Jiashi
  Feng, and Yannis Kalantidis.
\newblock Decoupling representation and classifier for long-tailed recognition.
\newblock In {\em Eighth International Conference on Learning Representations
  (ICLR)}, 2020.

\bibitem{Kornblith_2019_CVPR}
Simon Kornblith, Jonathon Shlens, and Quoc~V. Le.
\newblock Do better imagenet models transfer better?
\newblock In {\em Proceedings of the IEEE/CVF Conference on Computer Vision and
  Pattern Recognition (CVPR)}, June 2019.

\bibitem{scaling_od_classification_weights}
Jason Kuen, Federico Perazzi, Zhe Lin, Jianming Zhang, and Yap-Peng Tan.
\newblock Scaling object detection by transferring classification weights.
\newblock In {\em Proceedings of the IEEE/CVF International Conference on
  Computer Vision}, pages 6044--6053, 2019.

\bibitem{li2021improving}
Wei-Hong Li, Xialei Liu, and Hakan Bilen.
\newblock Improving task adaptation for cross-domain few-shot learning.
\newblock {\em CoRR}, 2021.

\bibitem{li2021rethinking}
Yang Li, Hong Zhang, and Yu Zhang.
\newblock Rethinking training from scratch for object detection, 2021.

\bibitem{Liang2015}
Xiaodan Liang, Si Liu, Yunchao Wei, Luoqi Liu, Liang Lin, and Shuicheng Yan.
\newblock Towards computational baby learning: A weakly-supervised approach for
  object detection.
\newblock In {\em 2015 IEEE International Conference on Computer Vision
  (ICCV)}, pages 999--1007, 2015.

\bibitem{fpn}
Tsung-Yi Lin, Piotr Doll{\'a}r, Ross Girshick, Kaiming He, Bharath Hariharan,
  and Serge Belongie.
\newblock Feature pyramid networks for object detection.
\newblock In {\em Proceedings of the IEEE conference on computer vision and
  pattern recognition}, pages 2117--2125, 2017.

\bibitem{retinanet}
Tsung-Yi Lin, Priya Goyal, Ross Girshick, Kaiming He, and Piotr Doll{\'a}r.
\newblock Focal loss for dense object detection.
\newblock In {\em Proceedings of the IEEE international conference on computer
  vision}, pages 2980--2988, 2017.

\bibitem{lin2014microsoft}
Tsung-Yi Lin, Michael Maire, Serge Belongie, Lubomir Bourdev, Ross Girshick,
  James Hays, Pietro Perona, Deva Ramanan, C.~Lawrence Zitnick, and Piotr
  Dollár.
\newblock Microsoft coco: Common objects in context, 2014.
\newblock cite arxiv:1405.0312Comment: 1) updated annotation pipeline
  description and figures; 2) added new section describing datasets splits; 3)
  updated author list.

\bibitem{liu2021swin}
Ze Liu, Yutong Lin, Yue Cao, Han Hu, Yixuan Wei, Zheng Zhang, Stephen Lin, and
  Baining Guo.
\newblock Swin transformer: Hierarchical vision transformer using shifted
  windows.
\newblock {\em arXiv preprint arXiv:2103.14030}, 2021.

\bibitem{Mahabadi21}
Rabeeh~Karimi Mahabadi, James Henderson, and Sebastian Ruder.
\newblock Compacter: Efficient low-rank hypercomplex adapter layers.
\newblock {\em CoRR}, abs/2106.04647, 2021.

\bibitem{mahajan2018exploring}
Dhruv Mahajan, Ross Girshick, Vignesh Ramanathan, Kaiming He, Manohar Paluri,
  Yixuan Li, Ashwin Bharambe, and Laurens Van Der~Maaten.
\newblock Exploring the limits of weakly supervised pretraining.
\newblock In {\em Proceedings of the European conference on computer vision
  (ECCV)}, pages 181--196, 2018.

\bibitem{Mahajan2018ExploringTL}
Dhruv Mahajan, Ross Girshick, Vignesh Ramanathan, Kaiming He, Manohar Paluri,
  Yixuan Li, Ashwin Bharambe, and Laurens van~der Maaten.
\newblock Exploring the limits of weakly supervised pretraining.
\newblock In Vittorio Ferrari, Martial Hebert, Cristian Sminchisescu, and Yair
  Weiss, editors, {\em Computer Vision -- ECCV 2018}, pages 185--201, Cham,
  2018. Springer International Publishing.

\bibitem{pan2009survey}
Sinno~Jialin Pan and Qiang Yang.
\newblock A survey on transfer learning.
\newblock {\em IEEE Transactions on knowledge and data engineering}, 2009.

\bibitem{Papadopoulos2017ExtremeCF}
Dim~P. Papadopoulos, Jasper R.~R. Uijlings, Frank Keller, and Vittorio Ferrari.
\newblock Extreme clicking for efficient object annotation.
\newblock {\em 2017 IEEE International Conference on Computer Vision (ICCV)},
  pages 4940--4949, 2017.

\bibitem{pfeiffer2021adapterfusion}
Jonas Pfeiffer, Aishwarya Kamath, Andreas Rücklé, Kyunghyun Cho, and Iryna
  Gurevych.
\newblock Adapterfusion: Non-destructive task composition for transfer
  learning, 2021.

\bibitem{Puigcerver2021ScalableTL}
Joan Puigcerver, Carlos Riquelme, Basil Mustafa, C{\'e}dric Renggli,
  Andr{\'e}~Susano Pinto, Sylvain Gelly, Daniel Keysers, and Neil Houlsby.
\newblock Scalable transfer learning with expert models.
\newblock {\em ArXiv}, abs/2009.13239, 2021.

\bibitem{predet}
Vignesh Ramanathan, Rui Wang, and Dhruv Mahajan.
\newblock Predet: Large-scale weakly supervised pre-training for detection.
\newblock In {\em Proceedings of the IEEE/CVF International Conference on
  Computer Vision}, pages 2865--2875, 2021.

\bibitem{RebuffiBV17}
Sylvestre{-}Alvise Rebuffi, Hakan Bilen, and Andrea Vedaldi.
\newblock Learning multiple visual domains with residual adapters.
\newblock {\em CoRR}, abs/1705.08045, 2017.

\bibitem{Rebuffi18}
Sylvestre-Alvise Rebuffi, Hakan Bilen, and Andrea Vedaldi.
\newblock Efficient parametrization of multi-domain deep neural networks.
\newblock In {\em IEEE Conference on Computer Vision and Pattern Recognition},
  2018.

\bibitem{fasterrcnn}
Shaoqing Ren, Kaiming He, Ross Girshick, and Jian Sun.
\newblock Faster r-cnn: Towards real-time object detection with region proposal
  networks.
\newblock {\em Advances in neural information processing systems}, 28:91--99,
  2015.

\bibitem{Rosenberg2005}
Chuck Rosenberg, Martial Hebert, and Henry Schneiderman.
\newblock Semi-supervised self-training of object detection models.
\newblock In {\em 2005 Seventh IEEE Workshops on Applications of Computer
  Vision (WACV/MOTION'05) - Volume 1}, volume~1, pages 29--36, 2005.

\bibitem{russakovsky2015imagenet}
Olga Russakovsky, Jia Deng, Hao Su, Jonathan Krause, Sanjeev Satheesh, Sean Ma,
  Zhiheng Huang, Andrej Karpathy, Aditya Khosla, Michael Bernstein, et~al.
\newblock Imagenet large scale visual recognition challenge.
\newblock {\em International journal of computer vision}, 115(3):211--252,
  2015.

\bibitem{Shen2020}
Zhiqiang Shen, Zhuang Liu, Jianguo Li, Yu-Gang Jiang, Yurong Chen, and
  Xiangyang Xue.
\newblock Object detection from scratch with deep supervision.
\newblock {\em IEEE transactions on pattern analysis and machine intelligence},
  42(2):398--412, 2019.

\bibitem{sohn2020detection}
Kihyuk Sohn, Zizhao Zhang, Chun-Liang Li, Han Zhang, Chen-Yu Lee, and Tomas
  Pfister.
\newblock A simple semi-supervised learning framework for object detection.
\newblock In {\em arXiv:2005.04757}, 2020.

\bibitem{SunShr17}
Chen Sun, Abhinav Shrivastava, Saurabh Singh, and Abhinav Gupta.
\newblock Revisiting unreasonable effectiveness of data in deep learning era.
\newblock In {\em ICCV}, pages 843--852. IEEE Computer Society, 2017.

\bibitem{tan20201st}
Jingru Tan, Gang Zhang, Hanming Deng, Changbao Wang, Lewei Lu, Quanquan Li, and
  Jifeng Dai.
\newblock 1st place solution of lvis challenge 2020: A good box is not a
  guarantee of a good mask.
\newblock {\em arXiv preprint arXiv:2009.01559}, 2020.

\bibitem{efficientnet}
Mingxing Tan and Quoc Le.
\newblock Efficientnet: Rethinking model scaling for convolutional neural
  networks.
\newblock In {\em International Conference on Machine Learning}, pages
  6105--6114. PMLR, 2019.

\bibitem{TanPL20}
Mingxing Tan, Ruoming Pang, and Quoc~V. Le.
\newblock Efficientdet: Scalable and efficient object detection.
\newblock In {\em 2020 {IEEE/CVF} Conference on Computer Vision and Pattern
  Recognition, {CVPR} 2020, Seattle, WA, USA, June 13-19, 2020}, pages
  10778--10787. Computer Vision Foundation / {IEEE}, 2020.

\bibitem{thompson2021deep}
Neil~C Thompson, Kristjan Greenewald, Keeheon Lee, and Gabriel~F Manso.
\newblock Deep learning's diminishing returns: The cost of improvement is
  becoming unsustainable.
\newblock {\em IEEE Spectrum}, 58(10):50--55, 2021.

\bibitem{xu2021end}
Mengde Xu, Zheng Zhang, Han Hu, Jianfeng Wang, Lijuan Wang, Fangyun Wei, Xiang
  Bai, and Zicheng Liu.
\newblock End-to-end semi-supervised object detection with soft teacher.
\newblock {\em Proceedings of the IEEE/CVF International Conference on Computer
  Vision (ICCV)}, 2021.

\bibitem{Barret2020}
Barret Zoph, Golnaz Ghiasi, Tsung-Yi Lin, Yin Cui, Hanxiao Liu, Ekin~D Cubuk,
  and Quoc~V Le.
\newblock Rethinking pre-training and self-training.
\newblock In {\em Advances in Neural Information Processing Systems}, pages
  3833--3845, 2020.

\end{thebibliography}
}

\newpage
\appendix

\section{\fpn based models: Disentangling Capacity}
\label{apd:capacity}
 
\begin{figure}[ht!]
    \centering
    \includegraphics[width=\linewidth]{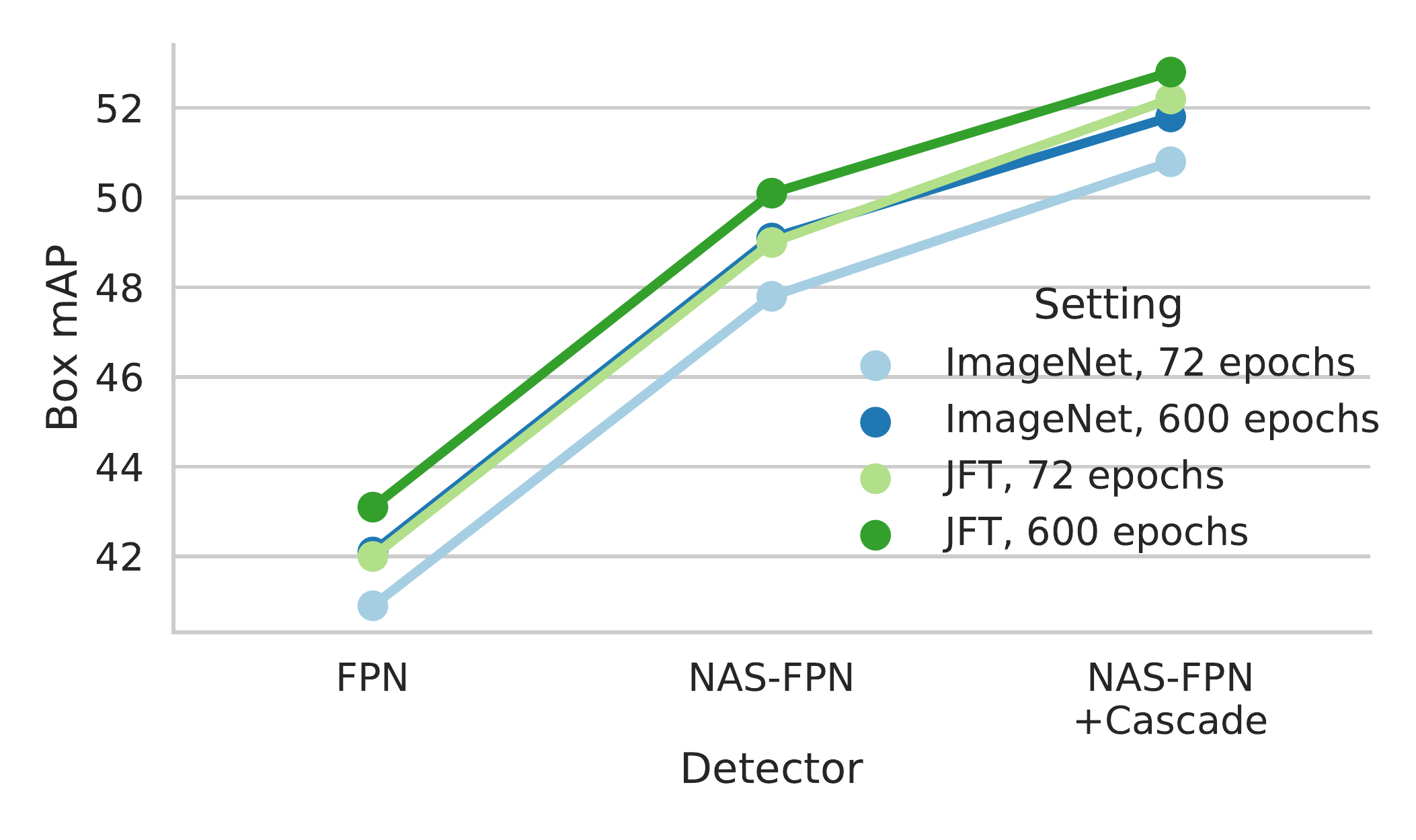}
    \caption{\label{fig:relative_improvement}
    Performance of different detectors preserving the same set of weights (\resnet-101). 
    As the backbones are frozen, it is possible to disentangle the relative benefit from pre-training on \Imagenet \emph{versus} \JFT from other confounding factors.
    Pre-training on a larger image classification dataset (\JFT) has a clear (and similar) benefit across detectors with different compositions (\fpn, \nasfpn, \nasfpn~+~\cascade). Training longer benefits all variations. 
}
\end{figure}

\begin{table}[ht!]
\centering
\scriptsize
\begin{tabular}{llrr}
  \toprule
    {\bf Model} & {\bf Pretraining} & \multicolumn{1}{r}{\bf mAP} & \multicolumn{1}{r}{\bf  AP $@$ 50} \\
  \midrule
  \fpn  \\
  \multirow{1}{*}{\makecell{\ \ \ +ResNet-50 }} & \Imagenet  &  $46.2$ & $68.5$ \\
     & \hspace{0.5em} + Freeze backbone & \mathdeltaneg{-6.5} $39.7$ & \mathdeltaneg{-6.2} $62.3$\\
     & \JFT  &  $46.4$  & $68.2$\\
     & \hspace{0.5em} + Freeze backbone & \mathdeltaneg{-6.0} $40.4$ & \mathdeltaneg{-4.1} $64.1$\\
    \multirow{1}{*}{\makecell{\ \ \ +ResNet-101}} & \Imagenet  &  $47.6$ & $69.0$ \\
     & \hspace{0.5em} + Freeze backbone & \mathdeltaneg{-6.7} $40.9$ & \mathdeltaneg{-5.0} $64.0$\\
     & \JFT  &  $48.1$  & $69.8$\\
     & \hspace{0.5em} + Freeze backbone & \mathdeltaneg{-6.1} $42.0$ & \mathdeltaneg{-3.6} $66.2$\\

  \midrule
  \fpn + \cascade \\
  \multirow{1}{*}{\makecell{\ \ \ +ResNet-50}} & \Imagenet  &  $48.5$ & $66.1$ \\
     & \hspace{0.5em} + Freeze backbone & \mathdeltaneg{-6.1} $42.4$ & \mathdeltaneg{-6.3} $59.8$\\
     & \JFT  &  $49.4$  & $67.3$\\
     & \hspace{0.5em} + Freeze backbone & \mathdeltaneg{-6.1} $43.3$ & \mathdeltaneg{-6.1} $61.4$\\
    \multirow{1}{*}{\makecell{\ \ \ +ResNet-101}}  & \Imagenet  &  $49.7$ & $67.8$ \\
     & \hspace{0.5em} + Freeze backbone & \mathdeltaneg{-5.5} $44.2$ & \mathdeltaneg{-5.8} $62.0$\\
     & \JFT  &  $50.3$  & $68.6$\\
     & \hspace{0.5em} + Freeze backbone & \mathdeltaneg{-5.1} $45.2$ &\mathdeltaneg{-4.6} $64.0$\\
\end{tabular}
\caption{\label{tab:fpn_resnet_shorter} Impact of freezing backbone on \fpn based models. Training for shorter (72 epochs) results. Models adopting \fpn do not benefit from feature preservation. The addition of \cascade heads do not change the observed results. The performance decrease is similar for models using \resnet-50 and \resnet-101 frozen backbones.}
\end{table}

\begin{table}[ht!]
\centering
\footnotesize
\begin{tabular}{lcrcr}
\toprule
{\bf Pretraining} \hspace{2pt} & {\bf conv. layers} 
&\multicolumn{1}{c}{\bf mAP} & &
\multicolumn{1}{c}{\bf  mAP $@$ 50} \\
\midrule
 \Scratch  
 & 1 &  $48.9 $  && $70.7$\\ 
 & 2 &  \mathdeltaneg{-0.5} $48.4 $  && \mathdeltaneg{-0.6} $70.1 $\\ 
 \midrule
  \multicolumn{4}{l}{ -- Fine-tune backbone}   \\
 \Imagenet  
 & 1 &$48.4 $  &&  $69.7$\\ 
 & 2 & \mathdelta{+0.2}  $ 48.6 $  && \mathdelta{+0.8} $70.5 $\\ 
 \JFT  
 & 1 & $ 48.6 $  &&  $ 70.1$\\ 
 & 2 &  \mathdelta{+0.1} $ 48.7 $  && \mathdelta{+0.4} $70.5 $\\ 
\midrule
  
 \multicolumn{4}{l}{ -- With frozen backbone}   
 \\
\Imagenet  
 & 1 &$41.3$  && $ 64.4$\\ 
& 2 &  \mathdelta{+0.8}  $42.1 $  && \mathdelta{+0.9} $65.3$\\ 
 \JFT   
 & 1 &   $42.2 $  && $66.4 $\\ 
 & 2 & \mathdelta{+0.9} $43.1 $  &&\mathdelta{+0.8} $67.2 $\\ 

\bottomrule
\end{tabular}
\caption{\label{tab:reduce_fc}
 Impact on \fpn based models performance  
 from decreasing their RPN capacity by reducing its convolutional layers from two (baseline value) to one layer. Results taken using a Resnet-101 backbone (training for longer). Training from scratch performance is increased with the capacity reduction while the opposite happens for models with pre-trained initialization. Models with frozen  backbone have the largest decrease in performance from reducing RPN's capacity.
}
\end{table}


\begin{table}[ht!]
\centering
\footnotesize
\begin{tabular}{lcrcr}
\toprule
{\bf Pretraining} \hspace{2pt} & {\bf conv. layers} 
&\multicolumn{1}{c}{\bf mAP} & &
\multicolumn{1}{c}{\bf  mAP $@$ 50} \\
\midrule

 \Scratch  & 2 &  $45.6$  && $67.0$\\ 
 & 4 &  \mathdelta{+0.3} $45.9$  && \mathdelta{+0.5} $67.5$\\ 
 \midrule
  \multicolumn{4}{l}{ -- Fine-tune backbone}   \\

 \Imagenet & 2 &  $47.5$  && $69.0$\\ 
 & 4 &  \mathdelta{+0.3} $47.8$  && \mathdelta{+0.7} $69.7$\\ 
 \JFT  & 2 &  $48.1$  && $69.8$\\ 
 & 4 &  \mathdeltaneg{-0.2} $47.9$  && \mathdelta{<0.1} $69.8$\\ 
\midrule
 \multicolumn{4}{l}{ -- With frozen backbone}   
 \\
\Imagenet & 2 &  $40.9 $  && $64.0$\\ 
 & 4 &  \mathdelta{+1.0} $41.9$  && \mathdelta{+1.2} $65.2$\\  
 \JFT &  2 &  $42.0$  && $66.2$\\ 
 & 4 &  \mathdelta{+0.9} $42.9$  && \mathdelta{+0.5} $66.7$\\ 
\bottomrule
\end{tabular}
\caption{\label{tab:rpn_four}
 Impact on  \fpn based models performance from augmenting RPN capacity by doubling its convolutional layers (from two to four layers). Results taken using a Resnet-101 backbone (training for 72 epochs). The gap between frozen and trained models remains large after increasing RPN's capacity.
}
\end{table}

\begin{table}
\centering
\footnotesize
\begin{tabular}{llrr}
  \toprule
 {\bf Pretraining} & {\bf $\#$filters} & \multicolumn{1}{r}{\bf mAP} & \multicolumn{1}{r}{\bf  AP $@$ 50} \\
  \midrule
  \multicolumn{4}{l}{ -- Full model fine-tuning}   \\
\Imagenet  & 256 & $47.6$ & $69.0$ \\
  & 512 & \mathdelta{+0.1}$47.7$ & \mathdelta{+0.1}$69.1$ \\
\JFT  & 256 & $48.1$  & $69.8$\\
 & 512 &  \mathdelta{<0.1} $48.1$ & \mathdelta{<0.1} $69.8$\\
\midrule
 \multicolumn{4}{l}{ -- With frozen backbone}  \\
 \hspace{0.5em} \Imagenet
& 256 & $40.9$ & $64.0$\\
& 512 & \mathdelta{+2.0} $42.9$ & \mathdelta{+1.9} $65.9$\\

\hspace{0.5em} \JFT & 256 & $42.0$ & $66.2$\\
& 512 & \mathdelta{+1.4} $43.4$ & \mathdelta{+1.0} $67.2$\\
  \midrule
\end{tabular}
\caption{\label{tab:resnet_512filters} 
Impact on \fpn based models performance from increasing detector components hidden representation form $256$ up to $512$.
Results  taken  using  a  \resnet-101  backbone (training for 72 epochs). Fine-tuned models show close to no improvement in performance. Models with frozen backbone benefit from extra capacity with a relative improvement. Their absolute performance, on the other hand, shows that extra capacity on filters alone is not enough to fully benefit from knowledge preservation.}
\end{table}

The experiments contained in this appendix further investigate the role of capacity in detectors adopting {\fpn}s.

As presented in the main text, knowledge preservation improves the performance of models with strong detector components (\nasfpn and \nasfpn+~\cascade) using both \resnet (\autoref{subsec:resnets}) and \efficientNet (\autoref{subsec:effic}) backbones.
Longer training schemes are able to change the backbone weights further away from a good initial representation,
but the relative increase in performance from  pre-training on a large dataset is clarified by comparing their frozen counterparts. 
As shown in \autoref{fig:relative_improvement}, the
the relative benefit from preserving the knowledge
from larger classification datasets is similar across different detectors. This is shown in the visualization as the lines comparing models pre-trained on \Imagenet are close to parallel to those of models pre-trained on \JFT. Individual performances can be found in \autoref{tab:resnet_longer}, \autoref{tab:fpn_resnet_shorter} and \autoref{tab:resnet_shorter}.

On the other hand, the absolute performances of models using {\fpn} show that those with frozen backbone lag behind their corresponding fine-tuned or trained from scratch (for longer) counterparts (\autoref{tab:fpn_resnet_shorter}). They also show that the addition of \cascade heads alone does not reduce the gap between the fine-tuned or frozen counterparts. 

The experiments in this appendix aim to disentangle the role of capacity from other confounding factors, in the gap observed on \fpn results. 
With that goal in mind, we ablate extra experiments that preserve the overall architecture of the \fpn based detector, but change the number of trainable parameters available.

First, we evaluate the impact of a small change in capacity, by decreasing the number of convolutional layers available on the RPN.  As \autoref{tab:reduce_fc} shows, this reduction in capacity impacts the performance of models with pre-trained backbones and those trained from scratch differently. While pre-trained ones have their performance reduced, the trained from scratch version benefits from it. The results also show that frozen models are more harmed by the decrease in capacity than the fine-tuned ones. 
\autoref{tab:rpn_four} presents the results of the opposite experiment, of increasing RPN's capacity by doubling the number of convolutional filters in this component. The increase in performance observed in frozen models is still small compared to the performance of fully trained counterparts. 

Next, we evaluate a larger change in number of parameters, by increasing the number of filters in the detector components from $256$ to $512$. More specifically we increase the hidden representation size on the RPN, Decoder and Detection head (see \autoref{fig:object_detection_diagram}). 
As \autoref{tab:resnet_512filters} shows, increasing capacity of the hidden representations does not impact fine-tuned models significantly, while improved the frozen counterpart. 

While the experiments explored in this section reduce the gap between the tuned and frozen \fpn models, they also show that changes in capacity alone do not fully explain the performance increase observed on stronger detectors. Next, we review the performance of \nasfpn based models.

\section{\nasfpn based models}
\label{apd:extra_forshorter}

\begin{table}
\centering
\scriptsize
\begin{tabular}{llrr}
  \toprule
    {\bf Model} & {\bf Pretraining} & \multicolumn{1}{r}{\bf mAP} & \multicolumn{1}{r}{\bf  AP $@$ 50} \\ 
     
  \midrule
  \nasfpn  \\
  \multirow{1}{*}{\makecell{\ \ \ +ResNet-50 }} & \Imagenet  &  $47.0$ & $68.0$ \\
     & \hspace{0.5em} + Freeze backbone & \mathdelta{+0.1} $47.1$ & \mathdelta{+0.3} $68.3$\\
     
     & \JFT  &  $47.5$  & $68.9$\\
     & \hspace{0.5em} + Freeze backbone & \mathdelta{+0.4} $47.9$ & \mathdelta{+0.7} $69.6$\\
    \multirow{1}{*}{\makecell{\ \ \ ResNet-101}} & \Imagenet  &  $48.2$ & $69.6$ \\
     & \hspace{0.5em} + Freeze backbone & \mathdeltaneg{-0.4} $47.8$ & \mathdeltaneg{-0.3} $69.3$\\
     & \JFT  & $48.5$ & $69.2$\\
     & \hspace{0.5em} + Freeze backbone & \mathdelta{+0.5} $49.0$ & \mathdelta{+1.3} $70.5$\\

  \midrule
  \nasfpn + \cascade \\
  \multirow{1}{*}{\makecell{\ \ \ +ResNet-50}} & \Imagenet  &  $49.4$ & $66.8$ \\
     & \hspace{0.5em} + Freeze backbone & \mathdelta{+0.5} $49.9$ & \mathdelta{+0.3} $67.1$\\
     & \JFT  &  $49.9$  & $67.6$\\
     & \hspace{0.5em} + Freeze backbone & \mathdelta{+1.1} $51.0$ & \mathdelta{+1.2} $68.8$\\
    \multirow{1}{*}{\makecell{\ \ \ +ResNet-101}} & \Imagenet  &  $51.1$ & $68.7  $ \\
     & \hspace{0.5em} + Freeze backbone & \mathdeltaneg{-0.3} $50.8$ & \mathdeltaneg{-0.3} $68.4$\\
     & \JFT  &  $50.9$  & $68.5$\\
     & \hspace{0.5em} + Freeze backbone & \mathdelta{+1.3} $52.2$ & \mathdelta{+1.5} $70.0$\\
  \bottomrule
\end{tabular}
\caption{\label{tab:resnet_shorter}  Impact on performance from frozen representation associated with the use of \nasfpn under shorter schedule regime (72 epochs). 
Freezing from \Imagenet takes longer to converge as the increase in accuracy observed on \nasfpn and \nasfpn+\cascade models is smaller than the observed in longer training regimes. Models composed with \nasfpn and frozen backbone produce matching or superior performance while consuming fewer resources during training.}
\end{table}

This section presents additional results with a shorter training schedule than the $600$ epochs schedule used in the main text. We aim to address the impact of training time on our results and review  \cite{SunShr17}'s observations in light of more recent tricks and practices for training object detectors. In this appendix specifically, we extend their observations to models using \nasfpn backbones, not covered by the original work. Adding to that, by the time \cite{SunShr17} was published, object detection batch size was considerably smaller than the $256$ size used in recent literature \cite{He2019RethinkingIP}. In \cite{SunShr17}, \MSCOCO training scheme adopted a batch size of $9$ images and a maximum of $3M$ steps. The only data augmentation used by them is random flipping, while recent findings show the importance of large scale jittering (LSJ)\cite{TanPL20} on detector training.
Thus, Tables \ref{tab:resnet_shorter} and \ref{tab:fpn_resnet_shorter} observe the impact of training for $72$ epochs with larger batch size ($64$) and the use of stronger data augmentation (LSJ).
We also note that \nasfpn was proposed after \cite{SunShr17}, thus, our results extend their observations to the use of more recent components.
In summary, by replicating the comparison using more recent findings, we confirm that \cite{SunShr17}'s findings about \fpn based models on shorter training scheme are still valid. At the same time, our results show that \emph{ the benefit from the re-use of the knowledge from large scale image classification datasets is dependent on the choice of the feature pyramid network architecture more than any of the other detector components}.

\section{\nasfpn Single Stage Detectors}
\label{apd:extra_forshorter}

\begin{table}[t]
\centering
\scriptsize
\begin{tabular}{llrr}
  \toprule
    {\bf Model} & {\bf Pretraining} & \multicolumn{1}{r}{\bf mAP} & \multicolumn{1}{r}{\bf  AP $@$ 50} \\
  \midrule
  \nasfpn + ResNet-101 
  & \Imagenet  &  $ 44.0 $ & $ 63.0 $ \\
     & \hspace{0.5em} + Freeze backbone & \mathdelta{+0.0} $ 44.0$ & \mathdelta{+0.2} $ 63.2 $\\
     & \JFT  &  $ 44.8 $  & $ 63.7 $\\
     & \hspace{0.5em} + Freeze backbone & \mathdelta{+0.8} $ 45.6$ & 
     \mathdelta{+1.5 } $ 65.3 $\\
  \midrule
  \fpn + ResNet-101
  & \Imagenet  &  $ 42.6 $ & $ 62.2 $ \\
     & \hspace{0.5em} + Freeze backbone & \mathdeltaneg{-7.3} $ 35.3 $ & \mathdeltaneg{-6.0} $ 56.2 $\\
     & \JFT  &  $43.7$  & $63.0$\\
     & \hspace{0.5em} + Freeze backbone & \mathdeltaneg{-8.4} $ 35.3  $ &\mathdeltaneg{-4.8} $ 58.2 $ \\
  \bottomrule
\end{tabular}
\caption{\label{tab:one_stage}
 One-stage detector performance (RetinaNet, 72  epochs). Similarly to the two-stage detectors, models with frozen features based on \nasfpn obtain similar performance e significantly reducing resources used during training. 
}
\end{table}

\begin{table}[t]
\centering
\scriptsize
\begin{tabular}{llrr}
  \toprule
    {\bf Model} & {\bf Pretraining} & \multicolumn{1}{r}{\bf mAP} & \multicolumn{1}{r}{\bf  AP $@$ 50} \\
  \midrule
  \nasfpn + ResNet-101 
    & From scratch  &  $ 42.6 $ & $ 61.7 $ \\
    & \Imagenet  &  $ 44.3 $ & $ 63.3 $ \\
     & \hspace{0.5em} + Freeze backbone & \mathdelta{+1.5} $45.8 $ & \mathdelta{+1.9} $ 65.2 $\\
     & \JFT  &  $ 44.0 $  & $ 62.8 $\\
     & \hspace{0.5em} + Freeze backbone & \mathdelta{+2.4} $ 46.4$ & 
     \mathdelta{+3.5} $ 66.3$\\
  \midrule
  \fpn + ResNet-101
  & From scratch  &  $ 43.5 $ & $ 63.2 $ \\
  & \Imagenet  &  $ 43.8 $ & $ 63.6 $ \\
     & \hspace{0.5em} + Freeze backbone & \mathdeltaneg{-7.6} $ 36.2 $ & \mathdeltaneg{-6.2} $ 57.4$\\
     & \JFT  &  $44.3$  & $63.7$\\
     & \hspace{0.5em} + Freeze backbone & \mathdeltaneg{-8.3} $ 36.0$ &\mathdeltaneg{-4.6} $ 59.1$ \\
  \bottomrule
\end{tabular}
\caption{\label{tab:one_stage600}
 One-stage detector performance (RetinaNet) under longer training (600  epochs) shows similar trends to the two-stage detectors results: models with Nas-FPN benefit from feature preservation. Training for longer improved the results and relative gain of feature preservation for models based on \nasfpn but did not help closing the performance gap on those based on \fpn.
}
\end{table}

This appendix complements the observations of the main text taken using two-stage detectors with ablations adopting single stage detectors, more specifically RetinaNet \cite{retinanet}.
Table \autoref{tab:one_stage} presents the results obtained by training RetinaNet detectors using a ResNet-101 backbones composed with \fpn and \nasfpn backbones trained for 72 epochs. Table \autoref{tab:one_stage600} presents the ablations using similar models but trained for 600 epochs. 
Similarly to the observations taken using two-stage detectors, single stage models using \nasfpn also benefit from feature preservation.
\begin{table*}[hb!]
\centering
\footnotesize
\begin{tabular}{lrrrrrrrr}
  \toprule
   LVIS  & \multicolumn{4}{c}{\bf Box} & \multicolumn{4}{c}{\bf Mask} \\
  \cmidrule(r){2-5} \cmidrule(l){6-9} 
    & {\bf mAP$_{\mathrm{\bf}}$}  & {\bf mAP$_{\mathrm{\bf r}}$} & {\bf mAP$_{\mathrm{\bf c}}$} & {\bf mAP$_{\mathrm{\bf f}}$} & {\bf mAP$_{\mathrm{\bf}}$}  & {\bf mAP$_{\mathrm{\bf r}}$} & {\bf mAP$_{\mathrm{\bf c}}$} & {\bf mAP$_{\mathrm{\bf f}}$}  \\
  \midrule
    \multicolumn{9}{l}{\bf First stage results: regular training} \\
    \cite{ghiasi2020simple}'s baseline &  35.0 & 
    12.7 & 34.0 & 45.9 & 32.2
    & 13.4& 32.2&40.4 \\
    \hspace{0.5em} + Freeze backbone & \mathdelta{+0.9} 35.9& \mathdelta{+0.1} 12.8 & \mathdelta{+1.3} 35.3  & \mathdelta{+0.9} 46.8 & \mathdelta{+1.3} 33.5 & \mathdelta{+0.0} 13.4 & \mathdelta{+2.0} 34.2 & \mathdelta{+1.1} 41.5 \\
    
  \midrule
  \multicolumn{9}{l}{\bf Second stage: tunes detection-classifier final layer using class-balanced loss} \\
  \cite{ghiasi2020simple}'s baseline &  
  37.6 & 23.2	&36.0& 45.7 & 34.9&24.6 & 34.2 & 40.3
  \\
    \hspace{0.5em} + Freeze backbone & \mathdelta{+1.8} 39.4 & \mathdelta{+1.2} 24.4 & \mathdelta{+2.7} 38.7  & \mathdelta{+1.1} 46.8 & \mathdelta{+2.3} 37.2 & \mathdelta{+2.0} 26.6 & \mathdelta{+3.3} 37.5 & \mathdelta{+1.3} 41.6 \\
  \bottomrule
\end{tabular}
\caption{\label{tab:lvis_num_annotations_noda} Performance using \efficientNet + \nasfpn per number of annotations groups.
Freezing the backbone matches or surpasses fine-tuning performance in all cases (both detection and segmentation)
. Balanced loss improves frozen backbone performance more than the fine-tuned one.
As opposed to fine-tuning \cite{ghiasi2020simple}, preserving features induces an increase in performance on rare ($\mathrm{mAP}_\mathrm{r}$) and common ($\mathrm{mAP}_\mathrm{c}$) classes, while still improving frequent ($\mathrm{mAP}_\mathrm{f}$) classes. 
Original first phase results are provided by the authors of \cite{ghiasi2020simple}. Results with Copy-Paste augmentations can be found in
 \autoref{tab:lvis_num_annotations}.}
\end{table*}

\begin{table*}[hb!]
\centering
\footnotesize
\begin{tabular}{lrrrrrrrr}
  \toprule
  LVIS   & \multicolumn{4}{c}{\bf Box} & \multicolumn{4}{c}{\bf Mask} \\
  \cmidrule(r){2-5} \cmidrule(l){6-9} 
    & {\bf mAP$_{\mathrm{\bf}}$}  & {\bf mAP$_{\mathrm{\bf s}}$} & {\bf mAP$_{\mathrm{\bf m}}$} & {\bf mAP$_{\mathrm{\bf l}}$} & {\bf mAP$_{\mathrm{\bf}}$}  & {\bf mAP$_{\mathrm{\bf s}}$} & {\bf mAP$_{\mathrm{\bf m}}$} & {\bf mAP$_{\mathrm{\bf l}}$}  \\
  \midrule
    \multicolumn{9}{l}{\bf First stage results: regular training} \\
    \cite{ghiasi2020simple}'s baseline &  
    35.0 & 28.5 & 43.8 & 50.3 & 32.2 & 24.4 & 42.2 & 49.0\\
    \hspace{0.5em} + Freeze backbone & 
    \mathdelta{+0.9} 35.9 &
    \mathdelta{+0.2} 28.7 & \mathdelta{+2.4} 46.2 & \mathdelta{+2.4} 52.7 & \mathdelta{+1.3} 33.5 & 
    \mathdelta{+0.4} 24.8 & \mathdelta{+2.3} 44.5 & \mathdelta{+2.8} 51.8 \\
  \midrule
  \multicolumn{9}{l}{\bf Second stage: tunes detection-classifier final layer using class-balanced loss} \\
    \cite{ghiasi2020simple}'s baseline  & 37.6& 30.8 & 46.9 & 53.4 & 34.9 & 26.4 & 45.3 & 52.2\\
    \hspace{0.5em} + Freeze backbone & 
    \mathdelta{+1.8} 39.4 & 
    \mathdelta{+0.6} 31.4 & \mathdelta{+2.9} 49.8  & \mathdelta{+4.0} 57.4 &  \mathdelta{+2.3} 37.2 &  
    \mathdelta{+0.6} 27.0 & \mathdelta{+3.3} 48.6  & \mathdelta{+4.5} 56.7 \\
  \bottomrule
\end{tabular}
\caption{\label{tab:lvis_object_size_noda} Performance using \efficientNet + \nasfpn per object size groups. Freezing the backbone surpasses fine-tuning performance for object of all sizes (both detection and segmentation). It has the strongest positive performance impact on large objects ($\mathrm{mAP}_\mathrm{l}$), then medium-sized objects ($\mathrm{mAP}_\mathrm{m}$), and finally small objects ($\mathrm{mAP}_\mathrm{s}$).
Balanced loss improves frozen backbone performance more than the fine-tuned one.
Original first phase results are provided by the authors of \cite{ghiasi2020simple}. Comparison results with Copy-Paste augmentations can be found in \autoref{tab:lvis_object_size}.}
\end{table*}

\section{LVIS: with no data augmentation}
\label{apd:extra_lvis}

This section presents results on the \LVIS dataset without the use of Copy-Paste\cite{ghiasi2020simple} augmentation. Results from  \autoref{tab:lvis_num_annotations_noda} show that preserving the features obtained on large classification datasets improves performance of objects with different numbers of annotations, while results from
\autoref{tab:lvis_object_size_noda} show that the benefit is also observed across objects of different sizes. The tables show positive impact for both detection and segmentation tasks. 

\section{Feature preservation and adaptation}
\label{apd:adapters}
 
\begin{table*}[ht!]
\centering
\footnotesize
\begin{tabular}{llllrrrrrr}
\toprule
{\bf Model} & {\bf Pre\-training}&  {\bf Epochs} & {\bf mAP} & $\delta_{Ft}$ & $\delta_{Fz}$ & {\bf mAP50} & $\delta_{Ft}$ & $\delta_{Fz}$ \\
\midrule
FPN  + $fbb$ + $ra$ 
& \Imagenet & 72  & 43.9& \mathdeltaneg{-3.7}&\mathdelta{+3.0}  & 66.8  &\mathdeltaneg{-2.2} & \mathdelta{+2.8}\\
& & 600 &  45.0 &\mathdeltaneg{-3.6} &\mathdelta{+2.9} &  67.8 &\mathdeltaneg{-2.6} & \mathdelta{+2.5}\\ 
& JFT & 72 &  46.0  & \mathdeltaneg{-2.2} &\mathdelta{+3.9} & 69.2 & \mathdeltaneg{-0.6}	& \mathdelta{+3.0}\\
&&600& 46.7  &\mathdeltaneg{-2.0} &\mathdelta{+3.6} &  69.9  &\mathdeltaneg{-0.6} & \mathdelta{+2.7}\\
\midrule
\nasfpn   + $fbb$ + $ra$ 
& \Imagenet & 72  & 48.6    &\mathdelta{+0.4} &\mathdelta{+0.8} & 70.0 & \mathdelta{+0.4} & \mathdelta{+0.7}\\ 
 &&600 &  49.9& \mathdelta{+0.9} & \mathdelta{+0.8} & 71.4 & \mathdelta{+1.4} & \mathdelta{+1.1}\\ 
& JFT & 72 &   49.8  & \mathdelta{+1.3} & \mathdelta{+0.8} &  71.5 &\mathdelta{+2.3} & \mathdelta{+1.0}\\ 
&&600& 51.1 &\mathdelta{+1.9} &\mathdelta{+1.0} &  73.1 &\mathdelta{+2.9} & \mathdelta{+1.3}\\ 

\midrule
\nasfpn, Cascade  + $fbb$ + $ra$ 
& \Imagenet & 72  & 
51.8  &\mathdelta{+0.7}& \mathdelta{+1.0}& 69.7 & \mathdelta{+1.0}&\mathdelta{+1.3}\\
 && 600  &
53.0 & \mathdelta{+1.9} & \mathdelta{+1.2}& 71.2  & \mathdelta{+2.0} &\mathdelta{+1.2}\\

& JFT & 72 &  53.0 & \mathdelta{+2.1} &\mathdelta{+0.8} & 71.2 &\mathdelta{+2.7} & \mathdelta{+1.2}\\ 

&& 600 & 53.6 &\mathdelta{+2.5}&\mathdelta{+0.9}&  72.0 &\mathdelta{+3.0} & \mathdelta{+1.0}\\
\bottomrule
\end{tabular}
\caption{\label{tab:resnet_residual_detectors}Residual adapters across detectors with increasing capacity and fixed backbone (training for shorter): Adapting feature backbones while preserving original knowledge by freezing the original \resnet weights. Columns show mAP and mAP@50 and their difference to their corresponding model trained with fine-tuned features ($\delta_{Ft}$) and frozen features ($\delta_{Fz}$). All models using Resnet-101. $fbb$: backbone frozen on classification features, $ra$: residual adapters.}
\end{table*}

\begin{table*}
\centering
\footnotesize
\begin{tabular}{lllrrrrrrrr}
\toprule
{\bf Model} &{\bf Pre\-training}& {\bf Epochs}& {\bf mAP} &  $\delta_{Ft}$ & $\delta_{Fz}$ & {\bf mAP50} &  $\delta_{Ft}$ & $\delta_{Fz}$ \\
\midrule

ResNet-50 & \Imagenet & 72&  50.5 & \mathdelta{+1.1} & \mathdelta{+0.7} & 68.3  & \mathdelta{+1.5}& \mathdelta{+1.2}\\ 
 &  &  600 &   51.4 &\mathdelta{+1.1}& \mathdelta{+0.3} &  69.4   &\mathdelta{+1.4} & \mathdelta{+0.3}\\ 

& JFT & 72& 51.7  & \mathdelta{+1.8} & \mathdelta{+0.7} & 69.8 & \mathdelta{+2.2}  & \mathdelta{+1.1}\\ 
&  & 600 & 52.8  &\mathdelta{+2.4} & \mathdelta{+0.7} & 71.2   &\mathdelta{+3.1} & \mathdelta{+0.8}\\ 

\midrule
ResNet-101  & \Imagenet & 72&  51.8 &\mathdelta{+0.7}& \mathdelta{+1.0}& 69.7 & \mathdelta{+1.0}&\mathdelta{+1.3}\\
&  &   600 &
53.0 &  \mathdelta{+1.9} & \mathdelta{+1.2}& 71.2 & \mathdelta{+2.0} &\mathdelta{+1.2}\\
	
& JFT &  72&  53.0 & \mathdelta{+2.1} &\mathdelta{+0.8} & 71.2  & \mathdelta{+2.7} & \mathdelta{+1.2}\\ 
& & 600 &  53.6  &\mathdelta{+2.5}&\mathdelta{+0.9}&  72.0 &\mathdelta{+3.0} & \mathdelta{+1.0}\\
	
\bottomrule
\end{tabular}
\caption{\label{tab:resnet_residual_backbones} Residual adapters over backbones in two sizes and same detector structure (training for longer): Adapting feature backbones while preserving original knowledge by freezing the original \resnet weights. Table shows results on Fast-RCNN combined with \nasfpn and Cascade heads. Columns show mAP and mAP@50 and their difference to the corresponding model with fine-tuned features ($\delta_{Ft}$) and frozen features ($\delta_{Fz}$). $fbb$: backbone frozen on classification features, $ra$: residual adapters.}
\end{table*}

In this appendix we extend the results presented in \autoref{subsec:adapters} to further explore the use of Residual Adapters~\cite{RebuffiBV17,Rebuffi18} as a mechanism to balance preserving knowledge obtained from the larger dataset and maintaining some amount of adaptability in the backbone while learning on the target task.
We present the performance delta with respect to full backbone fine-tuning ($\delta_{Ft}$) and backbone freezing ($\delta_{Fz}$) to better highlight the impact of residual adapters on results presented in \autoref{tab:resnet_longer}.

We ablate the use of Residual Adapters across detectors using different compositions (\fpn, \nasfpn, \nasfpn+\cascade) and a fixed backbone (\resnet-101).
As shown in~\autoref{tab:resnet_residual_detectors}, the use of residual adapters increased the performance of frozen models in all compositions (positive $\delta_{Fz}$).
Detectors adopting \fpn presented the largest gain from the use of residual adapters, but their absolute performance still lags behind their fine-tuned counterpart (negative $\delta_{Ft}$). 
 
The largest absolute performance is obtained with \nasfpn-based detectors (with gains over both the fine-tuned and frozen baselines), at the cost of increased computational  resource requirements.

Next,~\autoref{tab:resnet_residual_backbones}
presents our results on the effect of varying the backbone while fixing the detector components on the stronger detector composition explored (\nasfpn+\cascade). Using this detector composition, the gain obtained by the use of adapters is positive no matter if using the smaller (\resnet-50) or larger (\resnet-101) backbone.

\end{document}